\documentclass[10pt,twocolumn,letterpaper]{article}

\usepackage{cvpr}              % To produce the CAMERA-READY version
\newcommand{\impro}[1]{{\hspace{0.05cm}{\color[HTML]{32CB00}\textbf{(+#1)}}}}
\newcommand{\decli}[1]{{\hspace{0.05cm}{\color[HTML]{FF0000}\textbf{(-#1)}}}}

\newcommand{\myparagraph}[1]{\vspace{2pt}\noindent{\bf #1}}

\definecolor{cvprblue}{rgb}{0.21,0.49,0.74}
\usepackage[pagebackref,breaklinks,colorlinks,allcolors=cvprblue]{hyperref}

\def\paperID{414} % *** Enter the Paper ID here
\def\confName{CVPR}
\def\confYear{2025}

\usepackage{multicol}
\usepackage{multirow}
\usepackage{pifont}

\usepackage{colortbl}

\title{Number it: Temporal Grounding Videos like Flipping Manga}
\author{
Yongliang Wu$^{1,2,4}$\footnote[1]{} \footnote[2]{}
\quad \quad Xinting Hu$^{3}$\footnote[1]{}
\quad \quad Yuyang Sun$^{1,2}$
\quad Yizhou Zhou$^{4}$\footnote[3]{}\\ 
\quad Wenbo Zhu$^{5}$
\quad Fengyun Rao$^{4}$
\quad Bernt Schiele$^{3}$
\quad Xu Yang$^{1,2}$\footnote[4]{}\\ \\
\small $^{1}$Southeast University 
\\ \small $^{2}$Key Laboratory of New Generation Artificial Intelligence Technology and \\ \small Its Interdisciplinary Applications (Southeast University), Ministry of Education, China \\ \small \quad $^{3}$Max Planck Institute for Informatics, Saarland Informatics Campus, Germany  \\
\small $^{4}$WeChat Vision, Tencent Inc. \quad $^{5}$University of California, Berkeley\\
\small \texttt{yongliang0223@gmail.com xuyang\_palm@seu.edu.cn}
}

\begin{document}
\twocolumn[{
\renewcommand\twocolumn[1][]{#1}
\maketitle
\begin{center}
    \captionsetup{type=figure}
    \vspace{-10pt}
    \includegraphics[width=1\textwidth]{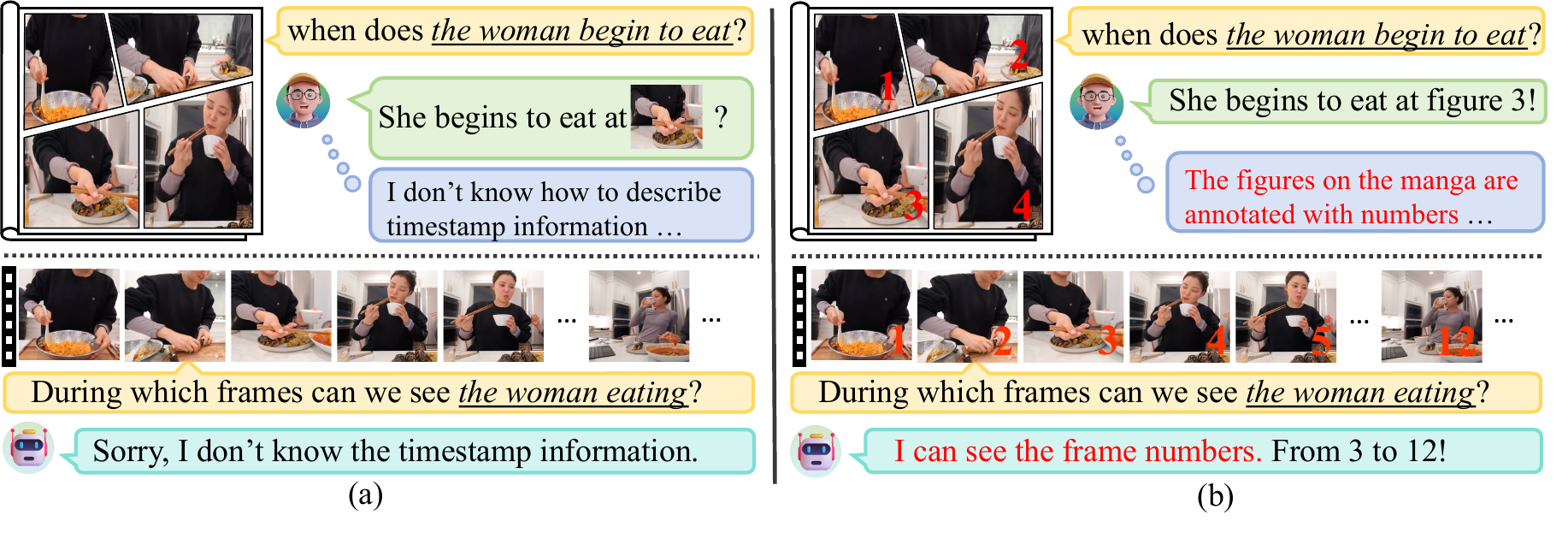}
    \vspace{-22pt}
    \captionof{figure}{\textbf{Effectiveness of Adding Frame Numbers for Temporal Grounding:} (a) Without numbered images or frames, both humans and Vid-LLMs struggle to locate specific timestamps accurately. (b) Once numbered,  grounding temporal cues becomes as intuitive as flipping manga, where timestamps are accessible at a glance.}
    \label{fig:teaser}
\end{center}
}]

\renewcommand{\thefootnote}{\fnsymbol{footnote}} 
\footnotetext{%
\footnotemark[1] Equal Contribution. \quad
\footnotemark[3] Project Leader. \quad 
\footnotemark[4] Corresponding Author.\\
\footnotemark[2] Work done during an internship at WeChat Vision, Tencent Inc. \quad
}
\begin{abstract}
Video Large Language Models (Vid-LLMs) have made remarkable advancements in comprehending video content for QA dialogue. However, they struggle to extend this visual understanding to tasks requiring precise temporal localization, known as Video Temporal Grounding (VTG). To address this, we introduce Number-Prompt (NumPro), a novel method that empowers Vid-LLMs to bridge visual comprehension with temporal grounding by adding unique numerical identifiers to each video frame. Treating a video as a sequence of numbered frame images, NumPro transforms VTG into an intuitive process: flipping through manga panels in sequence. This allows Vid-LLMs to ``read" event timelines, accurately linking visual content with corresponding temporal information. Our experiments demonstrate that NumPro significantly boosts VTG performance of top-tier Vid-LLMs without additional computational cost. Furthermore, fine-tuning on a NumPro-enhanced dataset defines a new state-of-the-art for VTG, surpassing previous top-performing methods by up to 6.9\% in mIoU for moment retrieval and 8.5\% in mAP for highlight detection. The code is available at \url{https://github.com/yongliang-wu/NumPro}.
\end{abstract}

% \vspace{0.1cm}
\section{Introduction}
\label{sec:intro}
Imagine you are watching a cooking video, and trying to locate the exact moment when the chef stirs in the spices. While recognizing such actions is feasible, translating that visual information into precise timing, i.e., a specific second or frame number, is surprisingly difficult. This challenge is central to the field of Video Temporal Grounding (VTG)~\cite{Charades,ANet,nan2021interventional,TimeChat,VTimeLLM,li2025llava}. In the realm of Video Large Language Models (Vid-LLMs)~\cite{peng2025lmm,Qwen2-VL,LLaVA,LLaVA-Video,LLaVA-OneVision,yi2024bridge} which process videos as a sequence of frame images, the integration of VTG allows for fine-grained visual and temporal understanding and reasoning of videos, which is pivotal for developing end-to-end video dialogue systems.

Despite advances of Vid-LLMs, endowing these models with effective VTG abilities presents a unique challenge: enhancing the model's visual recognition of an event within a video does not inherently enable it to describe when the event begins and ends using language~\cite{VTimeLLM,TimeChat}. For instance, advanced Vid-LLMs like Qwen2-VL~\cite{Qwen2-VL}, while excelling at video comprehension, can struggle with grounding specific events in time. When asked, e.g., to locate \textit{``when does the woman eat food"} in a 10-frame video, the model can hallucinate an illogical answer like \textit{``from frame 000 to 580."}\footnote{See more cases in Appendix~\ref{sec:general_vid}.} This limitation arises because these models are primarily trained to align visual content with language descriptions (what happens) while lacking mechanisms to directly interpret the temporal boundaries (when does it happen).
å

This gap in powerful Vid-LLMs leads us to think: How can we empower Vid-LLMs to extract temporal cues directly through visual recognition? A familiar human experience -- flipping manga -- provides an intuitive solution. When flipping manga, each numbered panel guides readers to follow the sequence of the narrative,  linking visual content with a clearly defined timeline. Inspired by this, we introduce Number-Prompt (NumPro), which places unique numerical identifiers on each video frame, similar to manga panel numbers.

% Current Vid-LLMs process videos by considering them as a sequence of frame images. 
With NumPro, VTG is as intuitive as flipping manga. 
As shown in Figure~\ref{fig:teaser}, NumPro augments each frame with a unique numerical identifier denoting its position in the temporal sequence. Given a language query targeting an event, Vid-LLMs retrieve relevant visual features of video frames and associate them with the frame numbers overlaid. These numerical identifiers are then directly translated into textual outputs. In practice, we strategically position frame numbers in the bottom-right corner, using a defined font size and distinct color. This design ensures numbers visibility without obstructing essential visual content. Overall, NumPro allows Vid-LLMs to ``read" the video timeline, effectively converting visual recognition into a temporal narrative.

NumPro's elegance lies in its simplicity: by subtly adding frame numbers as temporal markers into video frames, we enable Vid-LLMs to naturally correlate each frame to its temporal location in the video sequence. Unlike previous approaches~\cite{TimeChat,VTimeLLM,VTG-LLM,Hawkeye,GroundingGPT,Momentor,LITA}, NumPro does not introduce additional tokens or modify model vocabulary to provide temporal cues, thus avoiding additional learning complexities and maintaining strong transferability across various tasks and datasets. Temporal grounding, therefore, becomes an accessible, ``free-lunch” enhancement for Vid-LLMs already proficient in understanding video content. Additionally, fine-tuning on a specially curated NumPro-enhanced VTG dataset (NumPro-FT) further advances state-of-the-art performance.

Our contributions can be summarized as follows:
\begin{itemize}
    \item  We introduce NumPro, a novel approach that enhances Video Temporal Grounding (VTG) capabilities of Vid-LLMs by overlaying frame numbers onto video frames, making temporal grounding as intuitive as following numbered panels in flipping manga.
    \item Through an experimental study, we find a suitable NumPro design (font size, color, and position) that ensures high detectability by the model while minimally interfering with the original video content. 
    \item  We thoroughly evaluate NumPro on standard VTG benchmarks and metrics in both training-free and fine-tuned scenarios, demonstrating its effectiveness across various models and datasets. 
\end{itemize}

\begin{figure}[t]
    \centering
    \includegraphics[width=\linewidth]{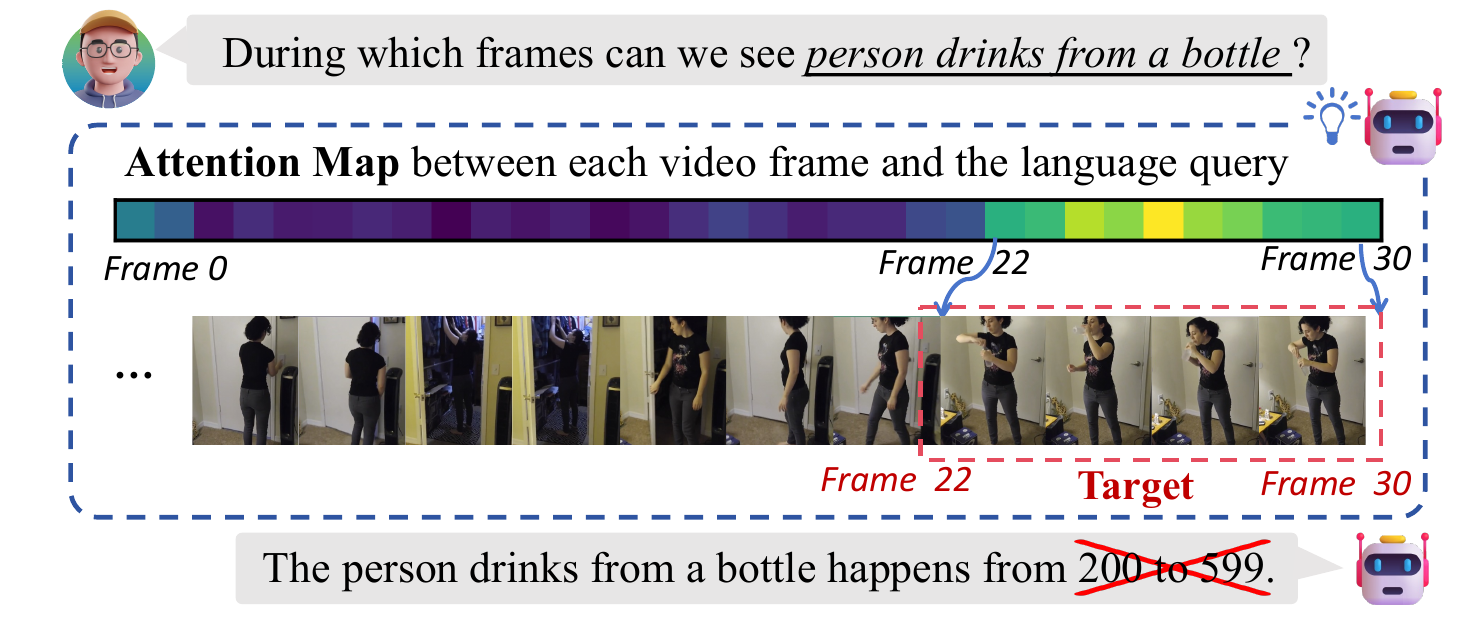}
    \caption{\textbf{Attention Analysis between Video Frames and Event Query.} Although the model accurately attends to regions of interest related to the query, it struggles to generate precise temporal boundaries in its response. }
    \label{fig:attn}
    \vspace{-10pt}
\end{figure}
\begin{figure*}[t]
    \centering
    \includegraphics[width=0.88\linewidth]{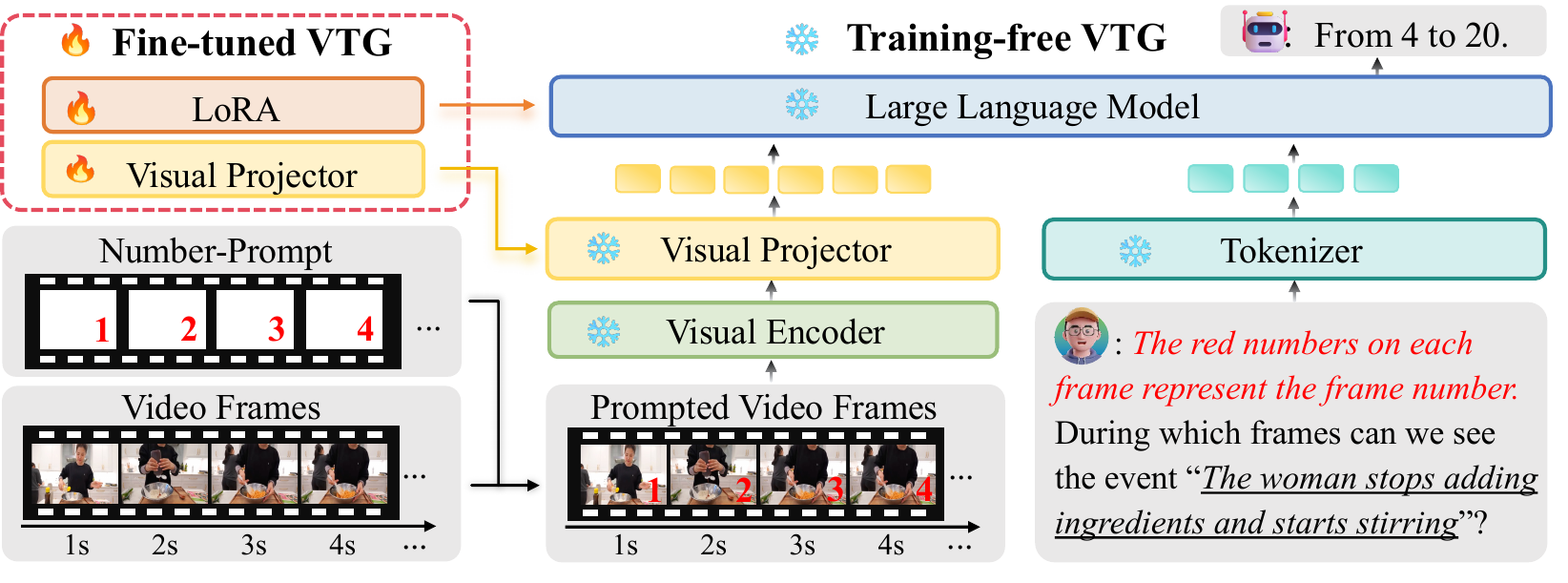}
    \caption{\textbf{Framework of Our Approach in Two Settings:} (1) Training-free VTG with NumPro, where frame numbers are directly added to video frames, enabling Vid-LLMs to locate events temporally without additional training, and (2) Fine-tuned VTG with NumPro-FT, which further improves VTG performance by fine-tuning Vid-LLMs on a dataset NumPro-enhanced with no architectural modifications.}
    \label{fig:method}
    \vspace{-10pt}
\end{figure*}

\section{Related Work}
\myparagraph{Video Temporal Grounding with Vid-LLMs.}
Video Temporal Grounding (VTG)~\cite{nan2021interventional,wang2018temporal,zhao2017temporal,li2020tea} focuses on the precise identification of event timestamps within videos, covering tasks such as moment retrieval~\cite{Charades,fastmr,mrblip,zala2023hierarchical,Grounded-VideoLLM,ANet,Image-text_retrieval,xu2022meta,foo2023system,xu2021sutd,liu2024generalized,ye2023vqne,ye2021visual,ye2022neighborhood,li2024frame}, dense captioning~\cite{krishna2017dense,kim2024you,vid2seq,Grounded-VideoLLM,ANet,person_search_survey,guo2024scaling,yin2024dataset,he2025homeomorphism,qian2024controllable,wang2023visual}, and highlight detection~\cite{QVHighlights,TimeChat}. 
For current Video Large Language Models (Vid-LLMs)~\cite{Qwen2-VL,LLaVA-Video,LLaVA-OneVision}, which leverage powerful LLMs~\cite{Qwen} for cross-modal understanding and video-based reasoning, VTG is crucial for achieving fine-grained temporal and visual comprehension, enabling end-to-end video dialogue systems with integrated temporal reasoning~\cite{VTimeLLM, TimeChat, Hawkeye}.
To achieve this, some methods rely on refined instruction datasets with temporal information (timestamps or frame numbers) for model fine-tuning~\cite{VTimeLLM, GroundingGPT}, while others concatenate additional textual temporal timestamps tokens with visual inputs~\cite{CogVLM2,Gemini} or introduce specific temporal embeddings~\cite{TimeChat, VTG-LLM}. Additional strategies model video structure~\cite{TRACE, LITA, Hawkeye} to better segment or organize videos into parts suitable for VTG. However, these approaches often require extensive retraining or specialized model adaptations, limiting their flexibility and transferability. In contrast, our NumPro aims to improve VTG for existing Vid-LLMs without additional training costs or architectural modifications.

\myparagraph{Visual Prompt in VLMs.} 
Visual prompts~\cite{vp_survey}, taking various forms such as circles~\cite{ViP-LLaVA,SoM}, bounding boxes~\cite{Cityllava, Instructdet, Groma} and semantic masks~\cite{CPT,ma2024invariant}, enhance vision-language models (VLMs)~\cite{ma2024visual,yang2024exploring,zhang2023pixel,wu2024glance, Add-it,OmniEdit} to focus on and reason about specific visual regions and reduce the occurrence of hallucination~\cite{ViP-LLaVA}. For CLIP~\cite{CLIP},  a simple red circle~\cite{RedCircle} or colored region~\cite{CPT} can effectively guide model attention. Multi-modal large language models (MLLMs)~\cite{Semantic-Aligned, Rasa, cao2020progressive}  are also sensitive to specific visual prompts~\cite{ViP-LLaVA}. For example, ViP-LLaVA~\cite{ViP-LLaVA} and SoM~\cite{SoM} prompt MLLMs to answer about specific image regions with graphic shapes or numeric tags. CoLLaVO~\cite{Collavo} and DOrA~\cite{DOrA} utilize pixel-level prompts in images or videos to enhance the semantic localization capability of MLLMs. Additionally, toolchain~\cite{Dettoolchain,Minedreamer,tziafas2024towards,sheng2024towards} approaches aggregate various visual prompts into multi-step reasoning paradigm to support reasoning complex tasks. While prior works focus on enhancing the region-based visual understanding of VLMs with visual prompts, our NumPro is the first to employ simple numerical tags as visual prompts within video frames to improve the temporal grounding capability.

\section{Number-Prompt Approach}
Our Number-Prompt (NumPro) approach provides a simple yet effective solution to enhance Video Temporal Grounding (VTG) capabilities of existing Video Large Language Models (Vid-LLMs) in both training-free and fine-tuned settings, as shown in Figure~\ref{fig:method}. Section~\ref{sec:attn} presents an attention analysis based on Qwen2-VL~\cite{Qwen2-VL} to highlight the challenge of aligning visual features with textual temporal boundaries. Section~\ref{sec:train} describes the construction of NumPro and the fine-tuning process of Vid-LLMs on a NumPro-augmented VTG dataset, referred to as NumPro-FT. Finally, Section~\ref{sec:design} details the design optimization of NumPro for maximizing its effectiveness.

\subsection{Attention Analysis}
\label{sec:attn}
Current Vid-LLMs process videos as a sequence of frames. Visual representations of the video can be taken as the concatenated representations from each individual frame, aggregating the information from discrete frames into a comprehensive video level. This allows Vid-LLMs to understand videos by aligning visual representations of frame images with the textual representations of language queries. 

To explore the challenge in video temporal grounding (VTG), we analyze the attention map between representations of the frame image tokens and the query language tokens, and then we assess the temporal description of relevant video frames. Using Qwen2-VL-7B~\cite{Qwen2-VL} as a case study, we highlight the challenge of VTG for Vid-LLMs: while Vid-LLMs can understand what event is happening within a video, they struggle to translate this understanding into a textual description that describes when the event begins and ends.

Specifically, we take a video and a language query as input, and extract the attention scores from the final multi-head self-attention layer of Qwen2-VL-7B~\cite{Qwen2-VL}. For each frame within the video sequence, we aggregate the attention scores from all the visual tokens corresponding to that frame across all attention heads. 
As illustrated in Figure~\ref{fig:attn}, the attention map reveals a strong correlation between the text query of an event and targeted video segments. It indicates that Qwen2-VL-7B can effectively focus on query-relevant frames, which is consistent with the model's strong performance in other content-related video understanding tasks~\cite{Video-MME,Mvbench,wu2024video}. However, the model struggles to verbalize the correct temporal boundaries, and generates surprising hallucinations such as ``from 200 to 599."\footnote{See more cases in Appendix~\ref{sec:attn_cases}}. This observation underscores the need for mechanisms that can bridge the gap between spatial feature alignment and temporal reasoning with Vid-LLMs, which we aim to address.

\subsection{NumPro and NumPro-FT}
\label{sec:train}
Our approach, Number-Prompt (NumPro), empowers Vid-LLMs to directly associate specific visual content with its temporal information, turning temporal localization into a visual alignment task. As shown in Figure~\ref{fig:method}, NumPro operates in both training-free and fine-tuned scenarios.

In the training-free setting, each video frame is marked with its corresponding frame number. By utilizing the built-in Optical Character Recognition (OCR) capabilities of Vid-LLMs, we enable them to ``read" the timeline through the frame numbers associated with visual content. To clarify the purpose of the added numbers to Vid-LLMs, we prepend a simple instruction to each event query: ``The red numbers on each frame represent the frame number." This approach allows Vid-LLMs to identify frame-level boundaries by directly linking the frame numbers to language queries.

For improved performance, NumPro-FT fine-tunes Vid-LLMs on a NumPro-augmented dataset. This stage aligns frame numbers with temporal spans within the training data, embedding temporal grounding capabilities into the model's learned representations. During fine-tuning, we freeze the visual encoder and only fine-tune the visual projector and LLM components. To reduce parameter count and training overhead, we apply Low-Rank Adaptation (LoRA)~\cite{LoRA} to adjust the LLM. Our training objective is to maximize the likelihood of generating the correct answer tokens \(\mathbf{A}\) via auto-regressive language modeling:
\begin{equation}
\begin{aligned}
P(\mathbf{A} \mid V, T_{\text{instruct}}) =
\prod_{j=1}^{L} P_\theta(A_j \mid V, X_{\text{instruct}}, \mathbf{A}_{<j})
\end{aligned}
\end{equation}

Here, \(V\) represents the input video, \(\theta\) denotes the trainable parameters, \(T_{\text{instruct}}\) is the text instruction, \(L\) is the length of the answer sequence \(\mathbf{A}\), and \(\mathbf{A}_{<j}\) includes all preceding answer tokens before the current token \(A_j\).

\subsection{Design of Numerical Prompt}
\label{sec:design}
An effective NumPro design must ensure: (1) numbers are easily recognized by the model, and (2) minimal interference with visual content. 
Previous research~\cite{ViP-LLaVA} indicates that the appearance and placement of visual prompts can influence model attention. Given that all Vid-LLMs operate at a fixed resolution of $336 \times 336$, we optimize NumPro by assessing three factors: font size, color, and placement position of the frame number.

To determine an effective NumPro design, we use two primary metrics: Number Accuracy, assessing how well the model identifies overlaid numbers, and Caption Accuracy, measuring how accurately the original caption aligns with frame content after adding numbers. Balancing these two metrics allows us to select NumPro configurations where the numbers are clearly recognizable without disrupting the main video content.

To make the design choices robust across various models and datasets, we employ CLIP-based experiments on a subset of MSCOCO dataset~\cite{COCO} to calculate Number Accuracy and Caption Accuracy separately. We use the CLIP ViT-B/32~\cite{CLIP,dosovitskiy2020image,luo2023zero,cao2024empirical,zhao2024multi,kong2024multi,jiang2024dg} model to generate visual and textual representations, as many Vid-LLMs utilize CLIP-style vision encoders~\cite{Qwen2-VL,Vlmevalkit,liu2024omniclip}, allowing our findings to generalize well across Vid-LLMs. COCO image-caption pairs serve as proxies for video frames, avoiding the high costs and limited scalability of direct VTG testing. Specifically, we randomly select 1,000 distinct image-caption pairs from MSCOCO~\cite{COCO} and overlay numbers ranging from ``0” to ``99” onto the image in various configurations.

\begin{figure}[tp]
    \centering
    \includegraphics[width=\linewidth]{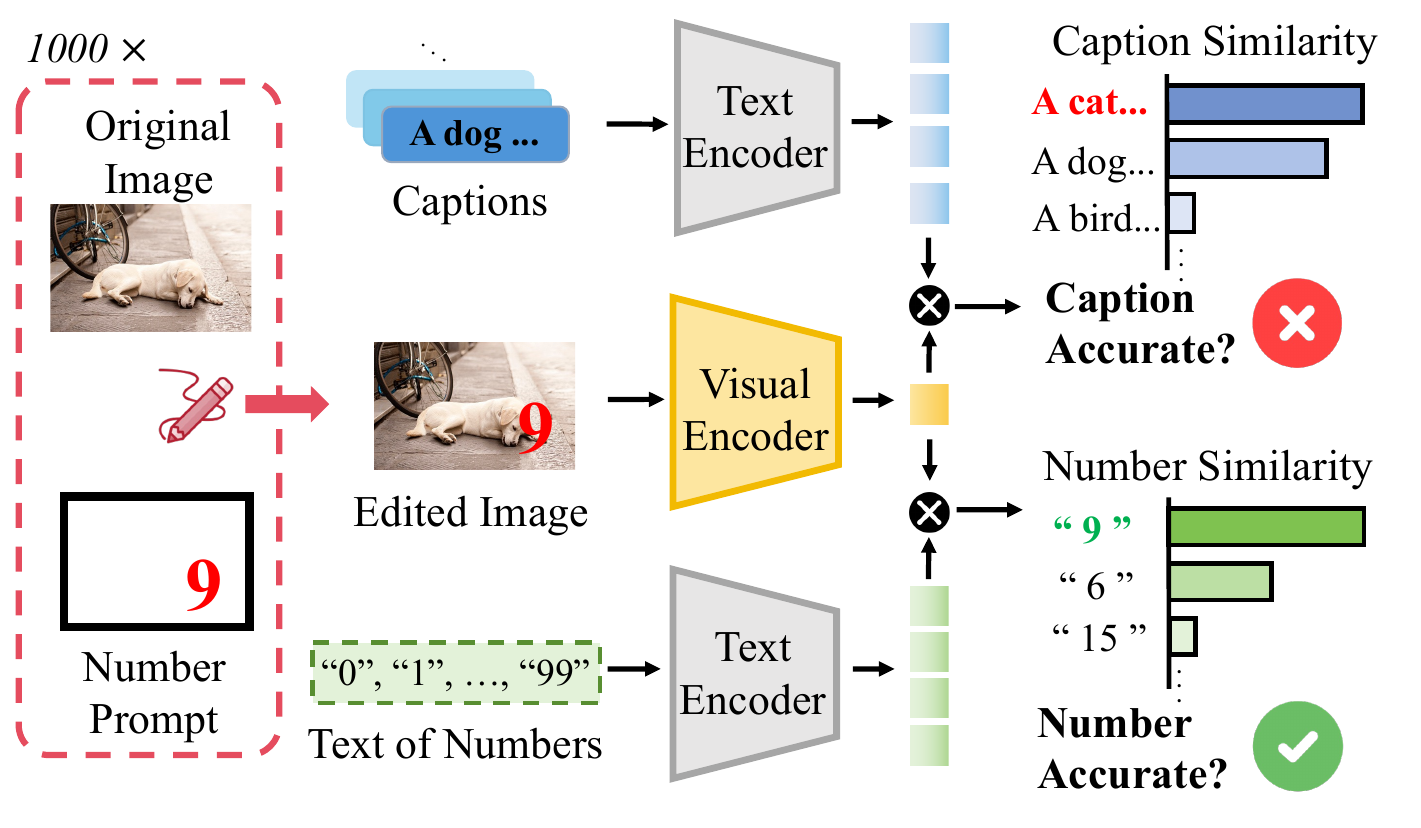}
    \caption{\textbf{ Our NumPro Design Algorithm.} We overlay different numbers onto COCO images and obtain visual and textual representations using CLIP encoders. For each configuration, we calculate Number/Caption Similarity and derive Number/Caption Accuracy, to identify the optimal NumPro design that balances recognizability and minimal disruption to the visual content. }
    \label{fig:clip}
    \vspace{-10pt}
\end{figure}

As shown in Figure~\ref{fig:clip}, we first obtain representations from CLIP~\cite{CLIP} vision and text encoders and compute intermediate similarity scores (\textit{i.e.,} Number and Caption Similarity) between them. Using the added numbers and original captions as ground truth, we select the text numbers and captions with the highest similarity scores as predictions to calculate Number and Caption Accuracy. Configurations balancing these accuracies are optimal for NumPro design.

\begin{figure}[t]
    \centering
    \includegraphics[width=1\linewidth]{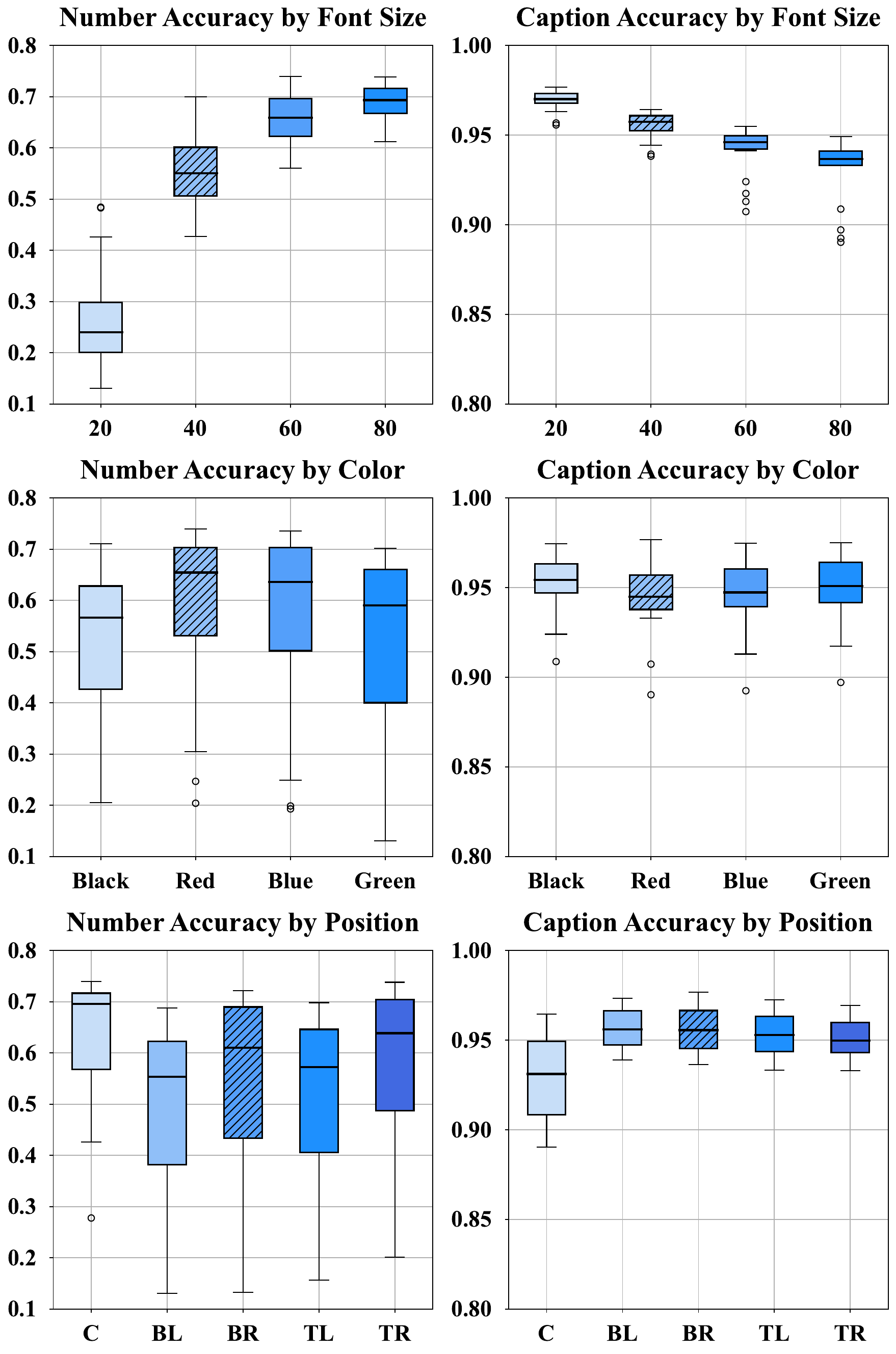}
    \caption{\textbf{The Impact of Different Number-Prompt Designs.} We categorize the design into three dimensions: font size, position, and color. BL stands for Bottom Left, BR for Bottom Right, TL for Top Left, TR for Top Right, and C for Center.}
    \label{fig:acc_analysis}
    \vspace{-10pt}
\end{figure}

As shown in Figure~\ref{fig:acc_analysis}, our findings indicate that increasing the font size improves number accuracy but reduces caption accuracy, suggesting that a moderate font size (40 or 60) is optimal. For color selection, caption accuracy remains relatively stable across different colors. Red shows the best performance for number accuracy, while black was the least effective. This finding is also consistent with previous works~\cite{RedCircle,ViP-LLaVA}. Additionally, positioning the text in the center of the image significantly reduced caption accuracy due to overlaps with key visual elements, while placing the numbers in the bottom-right corner provides the best balance between caption and number accuracy. Finally, we select a font size of 40, the color red, and the bottom-right position for our final NumPro design. 
% This design search allows NumPro to better harness the inherent OCR and visual-language alignment capabilities of Vid-LLMs to empower video temporal grounding. 

In practice, CLIP-based designs provide approximate rather than definitive guidance, further testing on Vid-LLMs with a VTG dataset may yield additional model-specific insights. In Sec~\ref{sec:ablation_design}, consistent results further validate the effectiveness of our design.

\section{Experiments}
\begin{table*}[t]
\renewcommand\arraystretch{0.95}
\centering
\caption{\textbf{Comparison of performance on the video temporal grounding task with previous state-of-the-art methods.} \textit{NumPro} refers to the use of number prompts for augmentation during inference, while \textit{NumPro-FT} indicates fine-tuning with the number prompt augmentation instruction dataset. The best results are highlighted in \textbf{bold}, and the second-best are \underline{underlined}.}
\resizebox{\textwidth}{!}{
\begin{tabular}{l|llll|llll|ll}
\toprule
\multirow{2}{*}{Model} & \multicolumn{4}{c|}{Charades-STA} & \multicolumn{4}{c|}{ActivityNet} & \multicolumn{2}{c}{QVHighlights}\\
                 & R@0.3 & R@0.5 & R@0.7 & mIoU & R@0.3 & R@0.5 & R@0.7 & mIoU & mAP & HIT@1\\ 
\midrule
\multicolumn{11}{c}{\textit{VTG-Tuned Vid-LLMs}} \\
\midrule
GroundingGPT~\cite{GroundingGPT} & - & 29.6 & 11.9 & - & - & - & - & - & - & -\\
LITA~\cite{LITA} & - & - & - & - & - & 25.9 & - & 28.6 &- & - \\
VTG-LLM~\cite{VTG-LLM} & 52.0 & 33.8 & 15.7 & - & - & - & - & - & 16.5 & 33.5 \\
TimeChat~\cite{TimeChat} & 47.7 & 22.9 & 12.5 & 30.6 & 30.2 & 16.9 & 8.2 & 21.8 & 14.5 & 23.9 \\
VTimeLLM~\cite{VTimeLLM} & 51.0 & 27.5 & 11.4 & 31.2 & 44.0 & 27.8 & 14.3 & 30.4 & - & - \\
Momentor~\cite{Momentor} & 42.9 & 23.0 & 12.4 & 29.3 & 42.6 & 26.6 & 11.6 & 28.5 & 7.6 & - \\
HawkEye~\cite{Hawkeye} & 50.6 & 31.4 & 14.5 & 33.7 & \underline{49.1} & 29.3 & 10.7 & 32.7 & - & - \\
\midrule
\multicolumn{11}{c}{\textit{General Vid-LLMs}} \\
\midrule
GPT-4o~\cite{GPT-4o} & 55.0 & 32.0 & 11.5 & 35.4 & 33.3 & 21.2 & 10.4 & 23.7 & \underline{39.5}
& \underline{68.7} \\
\rowcolor{gray!20} ~~~~~~\textit{+NumPro} & 57.1 & 35.5 & 13.5 & 37.6 & 45.5 & \underline{30.8} & \underline{18.4} & \underline{33.6} & \textbf{40.5} & \textbf{70.7} \\
\midrule
Qwen2-VL-7B~\cite{Qwen2-VL} & 8.7 & 5.4 & 2.4 & 7.9 & 17.0 & 9.4 & 3.9 & 12.5 & 21.5 & 42.2\\
\rowcolor{gray!20} ~~~~~~\textit{+NumPro} & \underline{60.}7 & \underline{36.8} & \underline{15.9} & \underline{38.5} & 44.2 & 26.4 & 14.4 & 31.3 & 23.6 & 43.4\\
\midrule
LongVA-7B-DPO~\cite{LongVA} &  22.6 & 10.1 & 2.2 & 14.6 & 11.8 & 5.3 & 1.9 & 8.2 &14.2& 20.4\\
\rowcolor{gray!20} ~~~~~~\textit{+NumPro} &  27.2 & 10.3 & 2.9 & 18.9 & 20.1 & 10.8 & 5.4 & 15.2 & 15.3 & 24.3\\
\rowcolor{gray!20} ~~~~~~\textit{+NumPro-FT} & \textbf{63.8} & \textbf{42.0} & \textbf{20.6} & \textbf{41.4} & \textbf{55.6} & \textbf{37.5} & \textbf{20.6} & \textbf{38.8} & 25.0 & 37.2 \\
\bottomrule
\end{tabular}}
\vspace{-10pt}
\label{table:main}
\end{table*}

We evaluate our model on two Video Temporal Grounding (VTG) tasks: Moment Retrieval~\cite{ANet,Charades} and Highlight Detection~\cite{QVHighlights}. Moment Retrieval, given a language query describing an event, identifies the specific start and end video frames of the event. We utilize Charades-STA~\cite{Charades} and ActivityNet~\cite{ANet} as evaluation datasets, following previous works~\cite{VTimeLLM,TimeChat,Momentor,Hawkeye}. Evaluation metrics include the mean Intersection over Union (mIoU) and recall@1 at various IoU thresholds $m$ (R@\(m\)), where \(m\) is set to \{0.3, 0.5, 0.7\} following previous work~\cite{VTimeLLM,TimeChat}.  For Highlight Detection, which aims to locate and rank video frames based on their relevance to the language query, we use QVHighlights~\cite{QVHighlights} for evaluation. Evaluation metrics include mean Average Precision (mAP) and HIT@1 (the hit ratio of the highest-scored clip), as in ~\cite{TimeChat,QVHighlights,VTG-LLM}. Please see Appendix~\ref{sec:more_cases} for more task examples.

\subsection{Implementation Details for NumPro-FT}
\myparagraph{Dataset Preparation.} Our temporal grounding dataset consists of 70k question-answer pairs from DiDeMo~\cite{DiDeMo} and ActivityNet Caption~\cite{ANet} datasets. Additionally, we incorporate data from Stage 2 and Stage 3 of VTimeLLM dataset~\cite{VTimeLLM}. After filtering out invalid videos, we obtain a comprehensive instruction dataset totaling 220k samples. Each video in our dataset is augmented with our NumPro method by overlaying frame numbers directly onto the video frames. The question-answer pairs follow a consistent template: questions are formatted as ``During which frames can we see \{query\}?” and answers are formatted as ``From \(x\) to \(y\)'', where $x$ and $y$ denote the start and end frame numbers of the query event.  

\myparagraph{Training Details.} We utilize the LongVA-7B-DPO~\cite{LongVA} as our base model, taking into account its uncomplicated design and its extensive capacity to handle context length. Additionally, it has not been trained on any video data. The model is trained for 3 epochs over our curated dataset with a total batch size of 128. We use the AdamW optimizer~\cite{Adam} with cosine learning rate decay. The learning rate is set to 1e-4, and the warm-up ratio is 0.05. The LLM component utilizes LoRA with parameters \(r = 64\) and \(\alpha = 128\). All experiments are conducted on 8 H800 GPUs.

\subsection{Main Results}
\subsubsection{Comparison with State-of-the-Art Methods}
\label{sec:quant}
Table~\ref{table:main} presents a comparative analysis of Vid-LLMs enhanced with our NumPro/NumPro-FT against existing state-of-the-art (SOTA) methods on Moment Retrieval and Highlight Detection tasks.

\myparagraph{Moment Retrieval:} Applying training-free NumPro enables Vid-LLMs to approach or exceed previous SOTA performance, benefiting both closed-source and open-source Vid-LLMs.
GPT-4o~\cite{GPT-4o} already exhibits strong moment retrieval capabilities, and our NumPro further enhances the performance. In particular, NumPro achieves a 9.9\% increase in mIoU on ActivityNet, surpassing the previous SOTA by 0.9\%. Qwen2-VL-7B performs poorly initially and also sees a significant improvement with NumPro, averaging a 24.7\% increase in mIoU across datasets.

Moreover, starting from a relatively low baseline on LongVA-7B-DPO~\cite{LongVA}, our fine-tuning approach, NumPro-FT, establishes new SOTA across all metrics. On Charades-STA, it surpasses previous SOTA by 11.8\%, 8.2\%, 4.9\%, 7.7\% (R@0.3, R@0.5, R@0.7, mIoU), and on ActivityNet, it surpasses previous SOTA by 6.5\%, 8.2\%, 6.3\%, 6.1\% (R@0.3, R@0.5, R@0.7, mIoU). These results demonstrate that NumPro and NumPro-FT can utilize the superior video understanding abilities of existing Vid-LLMs and significantly enhance their moment retrieval capabilities.

\myparagraph{Highlight Detection:}
In this task, models like GPT-4o~\cite{GPT-4o} and Qwen2-VL have already achieved state-of-the-art (SOTA) performance. However, our NumPro approach consistently enhances their performance, with an average increase of 1.55\% in mean Average Precision (mAP) and 1.6\% in the hit ratio of the highest-scored clip (HIT@1). Additionally, applying NumPro-FT enables LongVA-7B-DPO to surpass existing SOTA by a large margin (+8.5\% in mAP and +3.7\% in HIT@1). These findings suggest that NumPro and NumPro-FT, which can be easily appended to current Vid-LLMs, hold substantial potential for further advancing temporal reasoning capabilities.

\subsubsection{Effectiveness of NumPro across Vid-LLMs}

\begin{table*}[t]
\centering
\caption{\textbf{Performance of Applying NumPro to Various Vid-LLMs and Ablation Results on \textit{NumPro-FT}.}}
\resizebox{\textwidth}{!}{
\begin{tabular}{l|llll|llll}
\toprule
\multirow{2}{*}{Model} & \multicolumn{4}{c|}{Charades-STA} & \multicolumn{4}{c}{ActivityNet} \\
& R@0.3 & R@0.5 & R@0.7 & mIoU & R@0.3 & R@0.5 & R@0.7 & mIoU \\
\midrule
LLaVA-OneVision-7B~\cite{LLaVA-OneVision} & 22.3 & 7.9 & 2.1 & 15.9 & 7.1 & 3.1 & 1.1 & 6.1 \\
\rowcolor{gray!20} ~~~~~~\textit{+NumPro} & $42.9_\impro{20.6}$ & $19.4_\impro{11.5}$ & $6.6_\impro{4.5}$ & $28.1_\impro{12.2}$ & $14.4_\impro{7.3}$ & $7.9_\impro{4.8}$ & $3.8_\impro{2.7}$ & $11.3_\impro{5.2}$ \\
\midrule
LLaVA-Video-7B~\cite{LLaVA-Video} & 11.8 & 2.7 & 0.1 & 9.8 & 7.4 & 3.1 & 1.2 & 6.2 \\
\rowcolor{gray!20} ~~~~~~\textit{+NumPro} & $56.7_\impro{44.8}$ &$25.6_\impro{22.9}$ & $8.6_\impro{8.5}$ & $34.6_\impro{24.8}$ & $25.2_\impro{17.8}$ & $15.2_\impro{12.1}$ & $8.4_\impro{7.2}$ & $18.6_\impro{12.4}$ \\
\midrule
Qwen2-VL-72B~\cite{Qwen2-VL} & 0.0 & 0.0 & 0.0 & 0.2 & 1.0 & 0.6 & 0.3 & 1.0 \\
\rowcolor{gray!20} ~~~~~~\textit{+NumPro} & $25.8_\impro{25.8}$ & $9.9_\impro{9.9}$ & $3.0_\impro{3.0}$ & $17.4_\impro{17.2}$ & $35.5_\impro{34.5}$ & $21.4_\impro{20.8}$ & $11.0_\impro{10.7}$ & $25.5_\impro{24.5}$ \\
\midrule
LongVA-7B-DPO~\cite{LongVA} &  22.6 & 10.1 & 2.2 & 14.6 & 11.8 & 5.3 & 1.9 & 8.2 \\
~~~~~~\textit{+FT} & 62.0 & 41.6 & 19.9 & 40.2 & 41.8 & 25.7 & 13.7 & 29.0 \\
\rowcolor{gray!20} ~~~~~~\textit{+NumPro-FT} & $63.8_\impro{41.2}$ & $42.0_\impro{31.9}$ & $20.6_\impro{18.4}$ & $41.4_\impro{26.8}$ & $55.6_\impro{43.8}$ & $37.5_\impro{32.2}$ & $20.6_\impro{18.7}$ & $38.8_\impro{30.6}$ \\
\bottomrule
\end{tabular}
}
\vspace{-10pt}
\label{table:ablation}
\end{table*}
Beyond surpassing SOTA, Table~\ref{table:ablation} demonstrates the broad applicability and scalability of NumPro across various Vid-LLMs in Video Temporal Grounding. We apply NumPro to additional Vid-LLMs, including LLaVA-Video-7B~\cite{LLaVA-Video}, LLaVA-OneVision-7B~\cite{LLaVA-OneVision}, and Qwen2-VL-72B~\cite{Qwen2-VL}, and observe notable performance improvements, with average mIoU gains reaching up to 18.1\% on Charades and 14.0\% on ActivityNet. Moreover, we conduct fine-tuning experiments with and without NumPro-augmented data (indicated as \textit{+FT} in Table~\ref{table:ablation}). Results show that NumPro-FT consistently outperforms conventional fine-tuning, particularly on longer video datasets like ActivityNet, where it achieves a substantial 9.8\% gain in mIoU. Additional studies on NumPro’s effectiveness for QVHighlights are provided in Appendix~\ref{sec:ablation_qvh}. Those observations underscore the effectiveness of NumPro across models and highlight its superior impact when combined with fine-tuning. 

\subsubsection{Qualitative Results}
\label{sec:quali}
\begin{figure*}[t]
    \centering
    \includegraphics[width=0.94\linewidth]{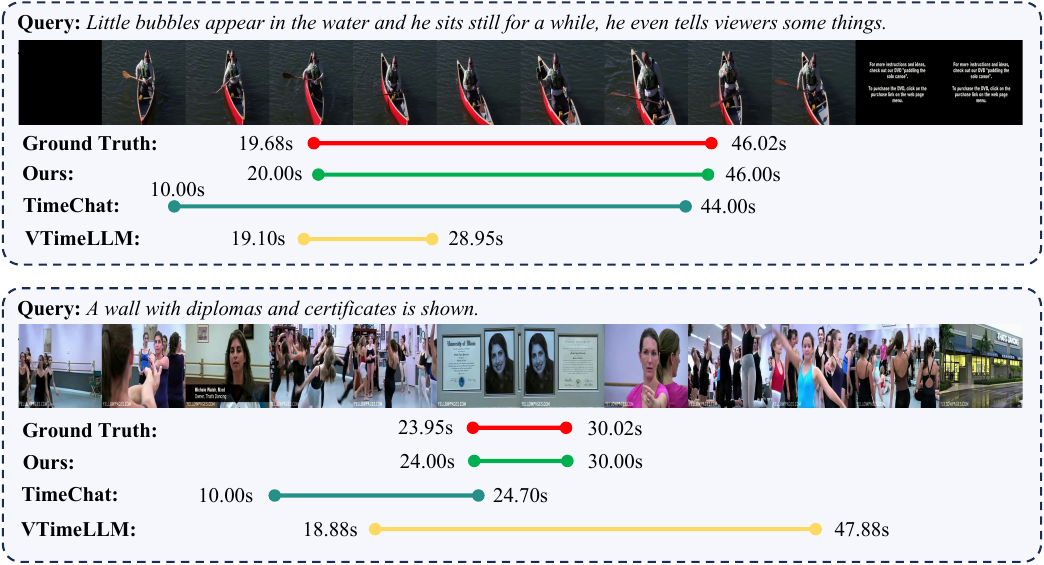}
    \caption{\textbf{Qualitative Comparison with State-of-the-Art.} Our LongVA-7B-DPO model, fine-tuned with NumPro-FT, outperforms TimeChat~\cite{TimeChat} and VTimeLLM~\cite{VTimeLLM} on ActivityNet by accurately identifying event boundaries in challenging scenes.}
    \vspace{-10pt}
    \label{fig:case}
\end{figure*}

In Figure~\ref{fig:case}, we compare our method with SOTA methods, TimeChat~\cite{TimeChat} and VTimeLLM~\cite{VTimeLLM}, through two visualization cases from our ActivityNet dataset.
The first example features minimal scene changes between video frames. TimeChat predicts an early start, while VTimeLLM fails to capture the full event duration. In contrast, our method precisely captures the correct event boundaries. The second case involves a shorter event duration and frequent scene changes. TimeChat completely misses the event, and VTimeLLM overestimates the event duration by including irrelevant segments. Our approach, again, precisely delineates the event boundaries. These qualitative examples underscore the robustness and precision of our method in scenarios that are especially challenging for other SOTA methods. We provide additional cases on moment retrieval and highlight detection in Appendix~\ref{sec:more_cases}.

\begin{table}[t]
\centering
\caption{\textbf{Ablation study on various NumPro designs.} We divide the designs into three dimensions: font size, color, and position.}
\resizebox{\linewidth}{!}{
\begin{tabular}{@{}lll|cccc@{}}
\toprule
\multirow{2}{*}{Size} & \multirow{2}{*}{Color} & \multirow{2}{*}{Position} & \multicolumn{4}{c}{Charades-STA}  \\
 &  &  & R@0.3 & R@0.5 & R@0.7 & mIoU \\ 
 \midrule
40 & Red & Top Left & 56.7 & 32.9 & 13.8 & 35.8 \\
40 & Red & Top Right & 58.2 & 34.0 & 13.0 & 36.8 \\
40 & Red & Center & 53.7 & 29.5 & 10.4 & 34.1 \\
40 & Red & Bottom Left & \textbf{61.6} & \textbf{37.8} & \textbf{15.9} & \textbf{39.3} \\
40 & Red & Bottom Right & \underline{60.7} & \underline{36.8} & \textbf{15.9} & \underline{38.5} \\
\midrule
20 & Red & Bottom Right & 53.6 & 34.0 & 14.0 & 34.6 \\
40 & Red & Bottom Right & \textbf{60.7} & \textbf{36.8} & \textbf{15.9} & \textbf{38.5} \\
60 & Red & Bottom Right &  \underline{58.0} &  \underline{34.5} &  \underline{14.1} &  \underline{37.1} \\
80 & Red & Bottom Right &  \underline{58.0} & 33.9 & 13.7 & 36.9 \\
\midrule
40 & Red & Bottom Right & \textbf{60.7} & \textbf{36.8} & \textbf{15.9} & \textbf{38.5} \\
40 & Blue & Bottom Right & \underline{57.8} & 34.2 & \underline{14.6} & \underline{36.6} \\
40 & Black & Bottom Right & 56.6 & \underline{36.0} & \textbf{15.9} & \underline{36.6} \\
40 & Green & Bottom Right & 56.0 & 33.8 & 14.5 & 36.0 \\
\bottomrule
\end{tabular}%
}
\vspace{-10pt}
\label{table:ablation_design}
\end{table}

\subsection{Validation of NumPro Design}
\label{sec:ablation_design}
Following our heuristic design process in Sec~\ref{sec:design}, we validate its effectiveness in temporal grounding tasks to confirm that these design choices generalize beyond the COCO dataset. We conduct moment retrieval experiments on Charades-STA~\cite{Charades} with Qwen2-VL-7B~\cite{Qwen2-VL} in a training-free setting. As shown in Table~\ref{table:ablation_design}, the results align closely with our initial observations from the COCO dataset, confirming the effectiveness of our design choices in VTG tasks. Specifically, (1) Position: Consistent with our CLIP-based findings, placing the text in the center has the largest impact on performance due to overlaps, while our choice of the bottom-right performs comparably to the best position; (2) Font Size: Both very large and very small fonts yield suboptimal results, supporting our balanced selection; (3) The performance on VTG is sensitive to number color, yet the red color consistently delivers the best performance, which may attribute to its high contrast against typical backgrounds~\cite{RedCircle}. Overall, the alignment between the CLIP-based design choices and the VTG results shows the validity and robustness of our NumPro design. Please refer to Appendix~\ref{sec:ablation_design_longva} for the ablation results of NumPro-FT. We also try directly overlaying timestamps (\textit{e.g.}, ``10.5s") on frames, which show inferior performance than frame numbers (Appendix~\ref{sec:timestamp_design}). 
\begin{figure}[t]
    \centering
    \includegraphics[width=\linewidth]{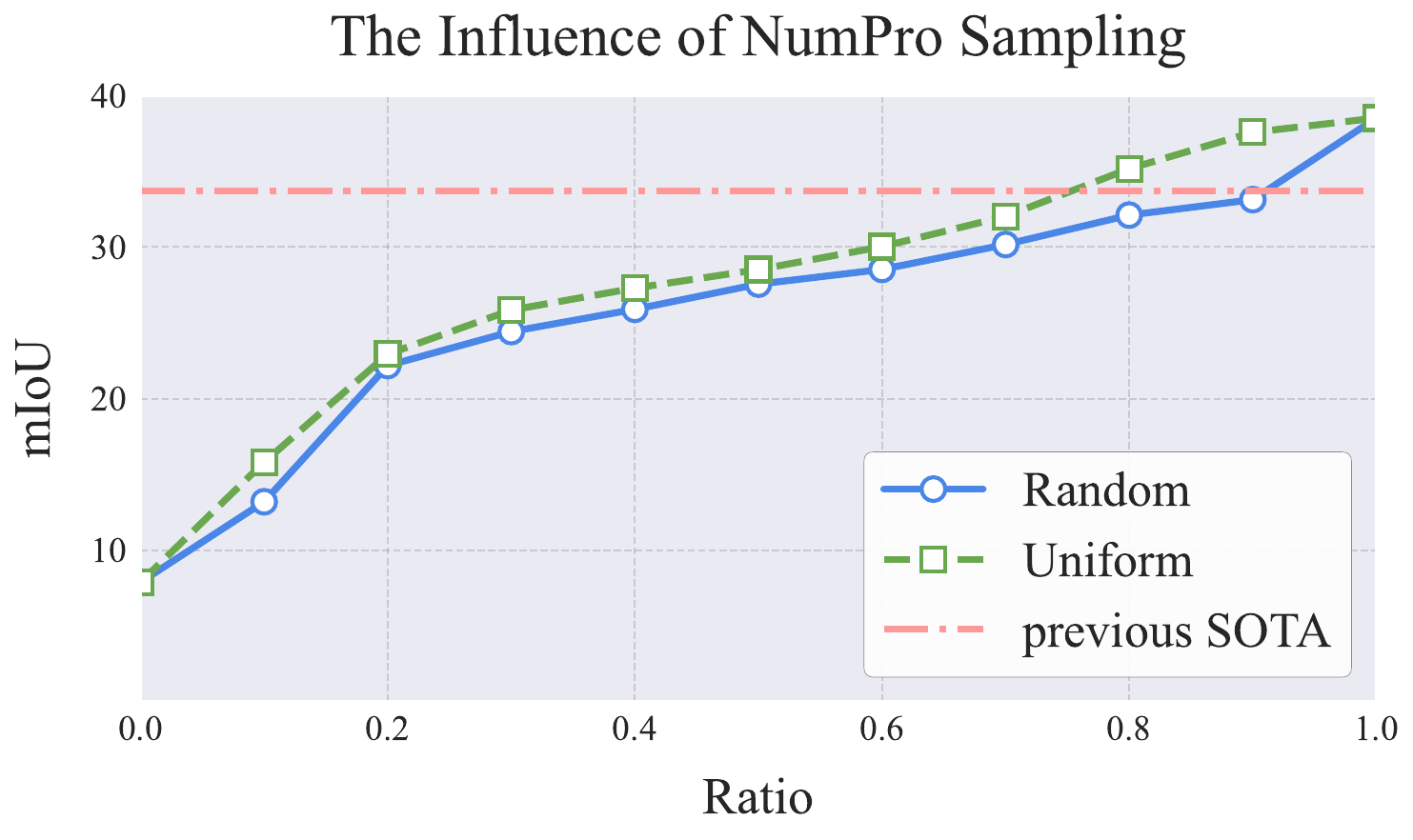}
    \caption{\textbf{Performance Comparison of Sampling Strategies for NumPro.} We compare the effects of NumPro with different sampling ratios and sampling methods (random vs. uniform), as tested on the Charades-STA~\cite{Charades} using the Qwen2-VL-7B~\cite{Qwen2-VL} model.}
    \vspace{-20pt}
    \label{fig:ratio}
\end{figure}

\subsection{Investigation on the Sampling of NumPro}
\label{sec:ablation_ratio}
Typically, we augment every frame in a video with NumPro. In this section, we evaluate the impact of varying the sampling ratio and sampling method (randomly or uniformly) when selecting a subset of frames from the video to augment NumPro. As depicted in Figure~\ref{fig:ratio}, performance increases with more labeled frames, with uniform sampling generally maintaining higher accuracy. Notably, labeling just 20\% of the frames provides a substantial performance boost and uniform sampling of 80\% of the frames surpasses previous state-of-the-art, underscoring the robustness of our NumPro approach.

\subsection{Influence on General Video-QA}
To explore the broader applicability of NumPro, we integrate it into general video-QA tasks, using VideoInstruct~\cite{Video-ChatGPT} as our benchmark. As detailed in Table~\ref{tab:video_qa}, the incorporation of NumPro minimally affects general comprehension metrics, with a slight decrease in Distraction Overlap (DO, -0.02) and an enhancement in Temporal Understanding (TU, +0.1). This indicates that Vid-LLMs equipped with NumPro maintain robust performance in general video-QA while excelling in precise video temporal grounding (VTG) tasks. This dual capability allows us to harness a powerful Vid-LLM for end-to-end video understanding that can flexibly adapt to both general and temporally nuanced questions within conversational AI systems. Moreover, we examine NumPro on more video-QA benchmarks including MVBench~\cite{Mvbench} and VideoMME~\cite{Video-MME}, and we show Vid-LLMs enhanced with NumPro achieve robust performance across a variety of downstream tasks. Details can be found in Appendix~\ref{sec:more_results_videoqa}.

\begin{table}[t]
\centering
\caption{\textbf{The influence of applying NumPro to general video-QA.} CI stands for correctness
of information, DO stands for detail orientation, CU stands for contextual understanding, TU stands for temporal understanding, and CO stands for consistency.}
\begin{tabular}{lccccc}
\toprule
Model & CI & DO & CU & TU & CO \\
\midrule
Qwen2-VL & 3.10 & 2.57 & 3.46 & 2.47 & 3.30 \\
\rowcolor{gray!20} ~~~~~~\textit{+NumPro} & 3.10 & 2.55 & 3.46 & 2.57 & 3.30 \\
\bottomrule
\end{tabular}
\vspace{-10pt}
\label{tab:video_qa}
\end{table}

\section{Conclusion}
In this paper, we propose Number-Prompt (NumPro), a simple yet efficient visual prompt designed to enhance the video temporal grounding (VTG) capabilities of Video Large Language Models (Vid-LLMs) with no effort. By overlaying frame numbers onto video content, NumPro leverages the inherent Optical Character Recognition (OCR) and visual-language alignment capabilities of Vid-LLMs, allowing them to accurately map events to specific temporal boundaries. Through systematic design informed by COCO-based heuristics and validated across VTG benchmarks, we demonstrated that NumPro effectively supports fine-grained temporal understanding while preserving general video comprehension. Through extensive evaluations, we demonstrated that NumPro consistently achieves state-of-the-art performance in both training-free and fine-tuned settings, enabling adaptable integration into both closed-source and open-source Vid-LLMs. NumPro-FT further refines temporal grounding performance, establishing new SOTA across VTG tasks. Besides, the minimal impact on general video-QA shows that NumPro can augment VTG while maintaining robust video understanding.

\section*{Acknowledgement}
This work is supported by the National Science Foundation of China (62206048), the Natural Science Foundation of Jiangsu Province (BK20220819), the Fundamental Research Funds for the Central Universities (2242024k30035), and the Southeast University Big Data Computing Center.
{
    \small
    \bibliographystyle{ieeenat_fullname}
    \bibliography{main}

\begin{thebibliography}{92}
\providecommand{\natexlab}[1]{#1}
\providecommand{\url}[1]{\texttt{#1}}
\expandafter\ifx\csname urlstyle\endcsname\relax
  \providecommand{\doi}[1]{doi: #1}\else
  \providecommand{\doi}{doi: \begingroup \urlstyle{rm}\Url}\fi

\bibitem[Bai et~al.(2023{\natexlab{a}})Bai, Bai, Chu, Cui, Dang, Deng, Fan, Ge, Han, Huang, et~al.]{Qwen}
Jinze Bai, Shuai Bai, Yunfei Chu, Zeyu Cui, Kai Dang, Xiaodong Deng, Yang Fan, Wenbin Ge, Yu Han, Fei Huang, et~al.
\newblock Qwen technical report.
\newblock \emph{arXiv preprint arXiv:2309.16609}, 2023{\natexlab{a}}.

\bibitem[Bai et~al.(2023{\natexlab{b}})Bai, Cao, Gao, Cao, Chen, Fan, Nie, and Zhang]{Rasa}
Yang Bai, Min Cao, Daming Gao, Ziqiang Cao, Chen Chen, Zhenfeng Fan, Liqiang Nie, and Min Zhang.
\newblock Rasa: Relation and sensitivity aware representation learning for text-based person search.
\newblock \emph{arXiv preprint arXiv:2305.13653}, 2023{\natexlab{b}}.

\bibitem[Boris et~al.(2024)Boris, Anil, Anna, and Marcus]{mrblip}
Meinardus Boris, Batra Anil, Rohrbach Anna, and Rohrbach Marcus.
\newblock The surprising effectiveness of multimodal large language models for video moment retrieval.
\newblock \emph{arXiv preprint arXiv:2406.18113}, 2024.

\bibitem[Caba~Heilbron et~al.(2015)Caba~Heilbron, Escorcia, Ghanem, and Carlos~Niebles]{ANet}
Fabian Caba~Heilbron, Victor Escorcia, Bernard Ghanem, and Juan Carlos~Niebles.
\newblock Activitynet: A large-scale video benchmark for human activity understanding.
\newblock In \emph{Proceedings of the ieee conference on computer vision and pattern recognition}, pages 961--970, 2015.

\bibitem[Cai et~al.(2024)Cai, Liu, Mustikovela, Meyer, Chai, Park, and Lee]{ViP-LLaVA}
Mu Cai, Haotian Liu, Siva~Karthik Mustikovela, Gregory~P Meyer, Yuning Chai, Dennis Park, and Yong~Jae Lee.
\newblock Vip-llava: Making large multimodal models understand arbitrary visual prompts.
\newblock In \emph{Proceedings of the IEEE/CVF Conference on Computer Vision and Pattern Recognition}, pages 12914--12923, 2024.

\bibitem[Cao et~al.(2020)Cao, Chen, Dou, Hu, Peng, and Kuijper]{cao2020progressive}
Min Cao, Chen Chen, Hao Dou, Xiyuan Hu, Silong Peng, and Arjan Kuijper.
\newblock Progressive bilateral-context driven model for post-processing person re-identification.
\newblock \emph{IEEE Transactions on Multimedia}, 23:\penalty0 1239--1251, 2020.

\bibitem[Cao et~al.(2022)Cao, Li, Li, Nie, and Zhang]{Image-text_retrieval}
Min Cao, Shiping Li, Juntao Li, Liqiang Nie, and Min Zhang.
\newblock Image-text retrieval: A survey on recent research and development.
\newblock \emph{arXiv preprint arXiv:2203.14713}, 2022.

\bibitem[Cao et~al.(2024{\natexlab{a}})Cao, Bai, Zeng, Ye, and Zhang]{cao2024empirical}
Min Cao, Yang Bai, Ziyin Zeng, Mang Ye, and Min Zhang.
\newblock An empirical study of clip for text-based person search.
\newblock In \emph{Proceedings of the AAAI Conference on Artificial Intelligence}, pages 465--473, 2024{\natexlab{a}}.

\bibitem[Cao et~al.(2024{\natexlab{b}})Cao, Bai, Zeng, Ye, and Zhang]{person_search_survey}
Min Cao, Yang Bai, Ziyin Zeng, Mang Ye, and Min Zhang.
\newblock An empirical study of clip for text-based person search.
\newblock In \emph{Proceedings of the AAAI Conference on Artificial Intelligence}, pages 465--473, 2024{\natexlab{b}}.

\bibitem[Chen et~al.(2021)Chen, Tsai, and Yang]{chen2021end}
Yi-Wen Chen, Yi-Hsuan Tsai, and Ming-Hsuan Yang.
\newblock End-to-end multi-modal video temporal grounding.
\newblock \emph{Advances in Neural Information Processing Systems}, 34:\penalty0 28442--28453, 2021.

\bibitem[Dang et~al.(2023)Dang, Feng, Zhang, Ge, Song, Gong, Liu, Chen, Zhu, Zhao, et~al.]{Instructdet}
Ronghao Dang, Jiangyan Feng, Haodong Zhang, Chongjian Ge, Lin Song, Lijun Gong, Chengju Liu, Qijun Chen, Feng Zhu, Rui Zhao, et~al.
\newblock Instructdet: Diversifying referring object detection with generalized instructions.
\newblock \emph{arXiv preprint arXiv:2310.05136}, 2023.

\bibitem[Dosovitskiy(2020)]{dosovitskiy2020image}
Alexey Dosovitskiy.
\newblock An image is worth 16x16 words: Transformers for image recognition at scale.
\newblock \emph{arXiv preprint arXiv:2010.11929}, 2020.

\bibitem[Duan et~al.(2024{\natexlab{a}})Duan, Yang, Qiao, Fang, Chen, Liu, Dong, Zang, Zhang, Wang, et~al.]{Vlmevalkit}
Haodong Duan, Junming Yang, Yuxuan Qiao, Xinyu Fang, Lin Chen, Yuan Liu, Xiaoyi Dong, Yuhang Zang, Pan Zhang, Jiaqi Wang, et~al.
\newblock Vlmevalkit: An open-source toolkit for evaluating large multi-modality models.
\newblock \emph{arXiv preprint arXiv:2407.11691}, 2024{\natexlab{a}}.

\bibitem[Duan et~al.(2024{\natexlab{b}})Duan, Cheng, Xu, Wu, Zhang, Ye, and Xie]{Cityllava}
Zhizhao Duan, Hao Cheng, Duo Xu, Xi Wu, Xiangxie Zhang, Xi Ye, and Zhen Xie.
\newblock Cityllava: Efficient fine-tuning for vlms in city scenario.
\newblock In \emph{Proceedings of the IEEE/CVF Conference on Computer Vision and Pattern Recognition}, pages 7180--7189, 2024{\natexlab{b}}.

\bibitem[Foo et~al.(2023)Foo, Gong, Fan, and Liu]{foo2023system}
Lin~Geng Foo, Jia Gong, Zhipeng Fan, and Jun Liu.
\newblock System-status-aware adaptive network for online streaming video understanding.
\newblock In \emph{Proceedings of the IEEE/CVF Conference on Computer Vision and Pattern Recognition}, pages 10514--10523, 2023.

\bibitem[Fu et~al.(2024)Fu, Dai, Luo, Li, Ren, Zhang, Wang, Zhou, Shen, Zhang, et~al.]{Video-MME}
Chaoyou Fu, Yuhan Dai, Yondong Luo, Lei Li, Shuhuai Ren, Renrui Zhang, Zihan Wang, Chenyu Zhou, Yunhang Shen, Mengdan Zhang, et~al.
\newblock Video-mme: The first-ever comprehensive evaluation benchmark of multi-modal llms in video analysis.
\newblock \emph{arXiv preprint arXiv:2405.21075}, 2024.

\bibitem[Gao and Xu(2021)]{fastmr}
Junyu Gao and Changsheng Xu.
\newblock Fast video moment retrieval.
\newblock In \emph{Proceedings of the IEEE/CVF International Conference on Computer Vision}, pages 1523--1532, 2021.

\bibitem[Gao et~al.(2017)Gao, Sun, Yang, and Nevatia]{Charades}
Jiyang Gao, Chen Sun, Zhenheng Yang, and Ram Nevatia.
\newblock Tall: Temporal activity localization via language query.
\newblock In \emph{Proceedings of the IEEE international conference on computer vision}, pages 5267--5275, 2017.

\bibitem[Guo et~al.(2024{\natexlab{a}})Guo, Wang, Zhang, Chin, Liu, Cheng, Pan, Lee, Xue, Shen, et~al.]{guo2024scaling}
Wei Guo, Hao Wang, Luankang Zhang, Jin~Yao Chin, Zhongzhou Liu, Kai Cheng, Qiushi Pan, Yi~Quan Lee, Wanqi Xue, Tingjia Shen, et~al.
\newblock Scaling new frontiers: Insights into large recommendation models.
\newblock \emph{arXiv preprint arXiv:2412.00714}, 2024{\natexlab{a}}.

\bibitem[Guo et~al.(2024{\natexlab{b}})Guo, Liu, Li, Tang, Chen, and Zhao]{VTG-LLM}
Yongxin Guo, Jingyu Liu, Mingda Li, Xiaoying Tang, Xi Chen, and Bo Zhao.
\newblock Vtg-llm: Integrating timestamp knowledge into video llms for enhanced video temporal grounding.
\newblock \emph{arXiv preprint arXiv:2405.13382}, 2024{\natexlab{b}}.

\bibitem[Guo et~al.(2024{\natexlab{c}})Guo, Liu, Li, Tang, Liu, and Chen]{TRACE}
Yongxin Guo, Jingyu Liu, Mingda Li, Xiaoying Tang, Qingbin Liu, and Xi Chen.
\newblock Trace: Temporal grounding video llm via causal event modeling.
\newblock \emph{arXiv preprint arXiv:2410.05643}, 2024{\natexlab{c}}.

\bibitem[He et~al.(2025)He, Wang, Ge, Chen, Yang, and Li]{he2025homeomorphism}
Yuting He, Boyu Wang, Rongjun Ge, Yang Chen, Guanyu Yang, and Shuo Li.
\newblock Homeomorphism prior for false positive and negative problem in medical image dense contrastive representation learning.
\newblock \emph{IEEE Transactions on Pattern Analysis and Machine Intelligence}, 2025.

\bibitem[Hong et~al.(2024)Hong, Wang, Ding, Yu, Lv, Wang, Cheng, Huang, Ji, Xue, et~al.]{CogVLM2}
Wenyi Hong, Weihan Wang, Ming Ding, Wenmeng Yu, Qingsong Lv, Yan Wang, Yean Cheng, Shiyu Huang, Junhui Ji, Zhao Xue, et~al.
\newblock Cogvlm2: Visual language models for image and video understanding.
\newblock \emph{arXiv preprint arXiv:2408.16500}, 2024.

\bibitem[Hu et~al.(2021)Hu, Wallis, Allen-Zhu, Li, Wang, Wang, Chen, et~al.]{LoRA}
Edward~J Hu, Phillip Wallis, Zeyuan Allen-Zhu, Yuanzhi Li, Shean Wang, Lu Wang, Weizhu Chen, et~al.
\newblock Lora: Low-rank adaptation of large language models.
\newblock In \emph{International Conference on Learning Representations}, 2021.

\bibitem[Huang et~al.(2024{\natexlab{a}})Huang, Wang, Chen, Song, and Zhu]{VTimeLLM}
Bin Huang, Xin Wang, Hong Chen, Zihan Song, and Wenwu Zhu.
\newblock Vtimellm: Empower llm to grasp video moments.
\newblock In \emph{Proceedings of the IEEE/CVF Conference on Computer Vision and Pattern Recognition}, pages 14271--14280, 2024{\natexlab{a}}.

\bibitem[Huang et~al.(2024{\natexlab{b}})Huang, Liao, Radhakrishnan, Yin, Molchanov, Yu, and Kautz]{LITA}
De-An Huang, Shijia Liao, Subhashree Radhakrishnan, Hongxu Yin, Pavlo Molchanov, Zhiding Yu, and Jan Kautz.
\newblock Lita: Language instructed temporal-localization assistant.
\newblock \emph{arXiv preprint arXiv:2403.19046}, 2024{\natexlab{b}}.

\bibitem[Jiang et~al.(2024)Jiang, Zhou, Li, Lu, Wang, Ma, Chang, and Zhang]{jiang2024dg}
Jincen Jiang, Qianyu Zhou, Yuhang Li, Xuequan Lu, Meili Wang, Lizhuang Ma, Jian Chang, and Jian~Jun Zhang.
\newblock Dg-pic: Domain generalized point-in-context learning for point cloud understanding.
\newblock In \emph{European Conference on Computer Vision}, pages 455--474. Springer, 2024.

\bibitem[Kim et~al.(2024)Kim, Kim, Moon, Choi, and Kim]{kim2024you}
Minkuk Kim, Hyeon~Bae Kim, Jinyoung Moon, Jinwoo Choi, and Seong~Tae Kim.
\newblock Do you remember? dense video captioning with cross-modal memory retrieval.
\newblock In \emph{Proceedings of the IEEE/CVF Conference on Computer Vision and Pattern Recognition}, pages 13894--13904, 2024.

\bibitem[Kingma(2014)]{Adam}
Diederik~P Kingma.
\newblock Adam: A method for stochastic optimization.
\newblock \emph{arXiv preprint arXiv:1412.6980}, 2014.

\bibitem[Kong et~al.(2024)Kong, Zhang, Wang, Zhou, Li, and Yuan]{kong2024multi}
Youyong Kong, Xiaotong Zhang, Wenhan Wang, Yue Zhou, Yueying Li, and Yonggui Yuan.
\newblock Multi-scale spatial-temporal attention networks for functional connectome classification.
\newblock \emph{IEEE Transactions on Medical Imaging}, 2024.

\bibitem[Krishna et~al.(2017)Krishna, Hata, Ren, Fei-Fei, and Carlos~Niebles]{krishna2017dense}
Ranjay Krishna, Kenji Hata, Frederic Ren, Li Fei-Fei, and Juan Carlos~Niebles.
\newblock Dense-captioning events in videos.
\newblock In \emph{Proceedings of the IEEE international conference on computer vision}, pages 706--715, 2017.

\bibitem[Lee et~al.(2024)Lee, Park, Kim, and Ro]{Collavo}
Byung-Kwan Lee, Beomchan Park, Chae~Won Kim, and Yong~Man Ro.
\newblock Collavo: Crayon large language and vision model.
\newblock \emph{arXiv preprint arXiv:2402.11248}, 2024.

\bibitem[Lei et~al.(2021)Lei, Berg, and Bansal]{QVHighlights}
Jie Lei, Tamara~L Berg, and Mohit Bansal.
\newblock Detecting moments and highlights in videos via natural language queries.
\newblock \emph{Advances in Neural Information Processing Systems}, 34:\penalty0 11846--11858, 2021.

\bibitem[Li et~al.(2024{\natexlab{a}})Li, Liu, Wang, and Yu]{li2024frame}
Bozheng Li, Mushui Liu, Gaoang Wang, and Yunlong Yu.
\newblock Frame order matters: A temporal sequence-aware model for few-shot action recognition.
\newblock \emph{arXiv preprint arXiv:2408.12475}, 2024{\natexlab{a}}.

\bibitem[Li et~al.(2024{\natexlab{b}})Li, Zhang, Guo, Zhang, Li, Zhang, Zhang, Li, Liu, and Li]{LLaVA-OneVision}
Bo Li, Yuanhan Zhang, Dong Guo, Renrui Zhang, Feng Li, Hao Zhang, Kaichen Zhang, Yanwei Li, Ziwei Liu, and Chunyuan Li.
\newblock Llava-onevision: Easy visual task transfer.
\newblock \emph{arXiv preprint arXiv:2408.03326}, 2024{\natexlab{b}}.

\bibitem[Li et~al.(2025)Li, Chen, Wei, Huang, Hui, Gao, Wei, and Liu]{li2025llava}
Hongyu Li, Jinyu Chen, Ziyu Wei, Shaofei Huang, Tianrui Hui, Jialin Gao, Xiaoming Wei, and Si Liu.
\newblock Llava-st: A multimodal large language model for fine-grained spatial-temporal understanding.
\newblock \emph{arXiv preprint arXiv:2501.08282}, 2025.

\bibitem[Li et~al.(2024{\natexlab{c}})Li, Wang, He, Li, Wang, Liu, Wang, Xu, Chen, Luo, et~al.]{Mvbench}
Kunchang Li, Yali Wang, Yinan He, Yizhuo Li, Yi Wang, Yi Liu, Zun Wang, Jilan Xu, Guo Chen, Ping Luo, et~al.
\newblock Mvbench: A comprehensive multi-modal video understanding benchmark.
\newblock In \emph{Proceedings of the IEEE/CVF Conference on Computer Vision and Pattern Recognition}, pages 22195--22206, 2024{\natexlab{c}}.

\bibitem[Li et~al.(2022)Li, Cao, and Zhang]{Semantic-Aligned}
Shiping Li, Min Cao, and Min Zhang.
\newblock Learning semantic-aligned feature representation for text-based person search.
\newblock In \emph{ICASSP 2022-2022 IEEE International Conference on Acoustics, Speech and Signal Processing (ICASSP)}, pages 2724--2728. IEEE, 2022.

\bibitem[Li et~al.(2020)Li, Ji, Shi, Zhang, Kang, and Wang]{li2020tea}
Yan Li, Bin Ji, Xintian Shi, Jianguo Zhang, Bin Kang, and Limin Wang.
\newblock Tea: Temporal excitation and aggregation for action recognition.
\newblock In \emph{Proceedings of the IEEE/CVF conference on computer vision and pattern recognition}, pages 909--918, 2020.

\bibitem[Li et~al.(2024{\natexlab{d}})Li, Xu, Zhang, Song, Cai, Qi, Zhou, Pan, Li, Vu, et~al.]{GroundingGPT}
Zhaowei Li, Qi Xu, Dong Zhang, Hang Song, Yiqing Cai, Qi Qi, Ran Zhou, Junting Pan, Zefeng Li, Van~Tu Vu, et~al.
\newblock Groundinggpt: Language enhanced multi-modal grounding model.
\newblock \emph{arXiv preprint arXiv:2401.06071}, 2024{\natexlab{d}}.

\bibitem[Lin et~al.(2023)Lin, Zhang, Chen, Pramanick, Gao, Wang, Yan, and Shou]{lin2023univtg}
Kevin~Qinghong Lin, Pengchuan Zhang, Joya Chen, Shraman Pramanick, Difei Gao, Alex~Jinpeng Wang, Rui Yan, and Mike~Zheng Shou.
\newblock Univtg: Towards unified video-language temporal grounding.
\newblock In \emph{Proceedings of the IEEE/CVF International Conference on Computer Vision}, pages 2794--2804, 2023.

\bibitem[Lin et~al.(2014)Lin, Maire, Belongie, Hays, Perona, Ramanan, Doll{\'a}r, and Zitnick]{COCO}
Tsung-Yi Lin, Michael Maire, Serge Belongie, James Hays, Pietro Perona, Deva Ramanan, Piotr Doll{\'a}r, and C~Lawrence Zitnick.
\newblock Microsoft coco: Common objects in context.
\newblock In \emph{Computer Vision--ECCV 2014: 13th European Conference, Zurich, Switzerland, September 6-12, 2014, Proceedings, Part V 13}, pages 740--755. Springer, 2014.

\bibitem[Liu et~al.(2024{\natexlab{a}})Liu, Li, Wu, and Lee]{LLaVA}
Haotian Liu, Chunyuan Li, Qingyang Wu, and Yong~Jae Lee.
\newblock Visual instruction tuning.
\newblock \emph{Advances in neural information processing systems}, 36, 2024{\natexlab{a}}.

\bibitem[Liu et~al.(2024{\natexlab{b}})Liu, Li, and Yu]{liu2024omniclip}
Mushui Liu, Bozheng Li, and Yunlong Yu.
\newblock Omniclip: Adapting clip for video recognition with spatial-temporal omni-scale feature learning.
\newblock \emph{arXiv preprint arXiv:2408.06158}, 2024{\natexlab{b}}.

\bibitem[Liu et~al.(2024{\natexlab{c}})Liu, Yang, Wang, Liu, Liu, Boukerche, Sun, and Song]{liu2024generalized}
Yang Liu, Dingkang Yang, Yan Wang, Jing Liu, Jun Liu, Azzedine Boukerche, Peng Sun, and Liang Song.
\newblock Generalized video anomaly event detection: Systematic taxonomy and comparison of deep models.
\newblock \emph{ACM Computing Surveys}, 56\penalty0 (7):\penalty0 1--38, 2024{\natexlab{c}}.

\bibitem[Luo et~al.(2023)Luo, Wang, Wu, Huang, and De~la Torre]{luo2023zero}
Jinqi Luo, Zhaoning Wang, Chen~Henry Wu, Dong Huang, and Fernando De~la Torre.
\newblock Zero-shot model diagnosis.
\newblock In \emph{Proceedings of the IEEE/CVF Conference on Computer Vision and Pattern Recognition}, pages 11631--11640, 2023.

\bibitem[Ma et~al.(2025)Ma, Jiang, Wu, Yuan, and Qi]{Groma}
Chuofan Ma, Yi Jiang, Jiannan Wu, Zehuan Yuan, and Xiaojuan Qi.
\newblock Groma: Localized visual tokenization for grounding multimodal large language models.
\newblock In \emph{European Conference on Computer Vision}, pages 417--435. Springer, 2025.

\bibitem[Ma et~al.(2024{\natexlab{a}})Ma, Xue, Wang, Zhou, Rao, Yan, Zhang, Wu, Shou, and Sun]{ma2024visual}
Feipeng Ma, Hongwei Xue, Guangting Wang, Yizhou Zhou, Fengyun Rao, Shilin Yan, Yueyi Zhang, Siying Wu, Mike~Zheng Shou, and Xiaoyan Sun.
\newblock Visual perception by large language model's weights.
\newblock \emph{arXiv preprint arXiv:2405.20339}, 2024{\natexlab{a}}.

\bibitem[Ma et~al.(2024{\natexlab{b}})Ma, Zhu, Zhang, Zhao, Wu, Huang, Hu, and Wu]{ma2024invariant}
Huan Ma, Yan Zhu, Changqing Zhang, Peilin Zhao, Baoyuan Wu, Long-Kai Huang, Qinghua Hu, and Bingzhe Wu.
\newblock Invariant test-time adaptation for vision-language model generalization.
\newblock \emph{arXiv preprint arXiv:2403.00376}, 2024{\natexlab{b}}.

\bibitem[Maaz et~al.(2023)Maaz, Rasheed, Khan, and Khan]{Video-ChatGPT}
Muhammad Maaz, Hanoona Rasheed, Salman Khan, and Fahad~Shahbaz Khan.
\newblock Video-chatgpt: Towards detailed video understanding via large vision and language models.
\newblock \emph{arXiv preprint arXiv:2306.05424}, 2023.

\bibitem[Mithun et~al.(2019)Mithun, Paul, and Roy-Chowdhury]{DiDeMo}
Niluthpol~Chowdhury Mithun, Sujoy Paul, and Amit~K Roy-Chowdhury.
\newblock Weakly supervised video moment retrieval from text queries.
\newblock In \emph{Proceedings of the IEEE/CVF Conference on Computer Vision and Pattern Recognition}, pages 11592--11601, 2019.

\bibitem[Nan et~al.(2021)Nan, Qiao, Xiao, Liu, Leng, Zhang, and Lu]{nan2021interventional}
Guoshun Nan, Rui Qiao, Yao Xiao, Jun Liu, Sicong Leng, Hao Zhang, and Wei Lu.
\newblock Interventional video grounding with dual contrastive learning.
\newblock In \emph{Proceedings of the IEEE/CVF conference on computer vision and pattern recognition}, pages 2765--2775, 2021.

\bibitem[OpenAI(2024)]{GPT-4o}
OpenAI.
\newblock Hello gpt-4o.
\newblock \url{https://openai.com/index/hello-gpt-4o/}, 2024.
\newblock Accessed: 2024-10-21.

\bibitem[Peng et~al.(2025)Peng, Zhang, Zhang, You, Liu, Zhu, Yang, Xu, Geng, and Yang]{peng2025lmm}
Yingzhe Peng, Gongrui Zhang, Miaosen Zhang, Zhiyuan You, Jie Liu, Qipeng Zhu, Kai Yang, Xingzhong Xu, Xin Geng, and Xu Yang.
\newblock Lmm-r1: Empowering 3b lmms with strong reasoning abilities through two-stage rule-based rl.
\newblock \emph{arXiv preprint arXiv:2503.07536}, 2025.

\bibitem[Qian et~al.(2024{\natexlab{a}})Qian, Li, Wu, Ye, Fei, Chua, Zhuang, and Tang]{Momentor}
Long Qian, Juncheng Li, Yu Wu, Yaobo Ye, Hao Fei, Tat-Seng Chua, Yueting Zhuang, and Siliang Tang.
\newblock Momentor: Advancing video large language model with fine-grained temporal reasoning.
\newblock \emph{arXiv preprint arXiv:2402.11435}, 2024{\natexlab{a}}.

\bibitem[Qian et~al.(2024{\natexlab{b}})Qian, Lin, See, and Li]{qian2024controllable}
Rui Qian, Weiyao Lin, John See, and Dian Li.
\newblock Controllable augmentations for video representation learning.
\newblock \emph{Visual Intelligence}, 2\penalty0 (1):\penalty0 1, 2024{\natexlab{b}}.

\bibitem[Radford et~al.(2021)Radford, Kim, Hallacy, Ramesh, Goh, Agarwal, Sastry, Askell, Mishkin, Clark, et~al.]{CLIP}
Alec Radford, Jong~Wook Kim, Chris Hallacy, Aditya Ramesh, Gabriel Goh, Sandhini Agarwal, Girish Sastry, Amanda Askell, Pamela Mishkin, Jack Clark, et~al.
\newblock Learning transferable visual models from natural language supervision.
\newblock In \emph{International conference on machine learning}, pages 8748--8763. PMLR, 2021.

\bibitem[Ren et~al.(2024)Ren, Yao, Li, Sun, and Hou]{TimeChat}
Shuhuai Ren, Linli Yao, Shicheng Li, Xu Sun, and Lu Hou.
\newblock Timechat: A time-sensitive multimodal large language model for long video understanding.
\newblock In \emph{Proceedings of the IEEE/CVF Conference on Computer Vision and Pattern Recognition}, pages 14313--14323, 2024.

\bibitem[Sheng et~al.(2024)Sheng, Chen, Tan, Liu, Chu, Bao, Gong, Liu, Xu, and Yu]{sheng2024towards}
Dianmo Sheng, Dongdong Chen, Zhentao Tan, Qiankun Liu, Qi Chu, Jianmin Bao, Tao Gong, Bin Liu, Shengwei Xu, and Nenghai Yu.
\newblock Towards more unified in-context visual understanding.
\newblock In \emph{Proceedings of the IEEE/CVF Conference on Computer Vision and Pattern Recognition}, pages 13362--13372, 2024.

\bibitem[Shtedritski et~al.(2023)Shtedritski, Rupprecht, and Vedaldi]{RedCircle}
Aleksandar Shtedritski, Christian Rupprecht, and Andrea Vedaldi.
\newblock What does clip know about a red circle? visual prompt engineering for vlms.
\newblock In \emph{Proceedings of the IEEE/CVF International Conference on Computer Vision (ICCV)}, pages 11987--11997, 2023.

\bibitem[Team et~al.(2023)Team, Anil, Borgeaud, Wu, Alayrac, Yu, Soricut, Schalkwyk, Dai, Hauth, et~al.]{Gemini}
Gemini Team, Rohan Anil, Sebastian Borgeaud, Yonghui Wu, Jean-Baptiste Alayrac, Jiahui Yu, Radu Soricut, Johan Schalkwyk, Andrew~M Dai, Anja Hauth, et~al.
\newblock Gemini: a family of highly capable multimodal models.
\newblock \emph{arXiv preprint arXiv:2312.11805}, 2023.

\bibitem[Tewel et~al.(2024)Tewel, Gal, Samuel, Atzmon, Wolf, and Chechik]{Add-it}
Yoad Tewel, Rinon Gal, Dvir Samuel, Yuval Atzmon, Lior Wolf, and Gal Chechik.
\newblock Add-it: Training-free object insertion in images with pretrained diffusion models, 2024.

\bibitem[Tziafas and Kasaei(2024)]{tziafas2024towards}
Georgios Tziafas and Hamidreza Kasaei.
\newblock Towards open-world grasping with large vision-language models.
\newblock \emph{arXiv preprint arXiv:2406.18722}, 2024.

\bibitem[Wang et~al.(2024{\natexlab{a}})Wang, Xu, Cheng, Diao, Zhou, Cao, Wang, Ge, and Huang]{Grounded-VideoLLM}
Haibo Wang, Zhiyang Xu, Yu Cheng, Shizhe Diao, Yufan Zhou, Yixin Cao, Qifan Wang, Weifeng Ge, and Lifu Huang.
\newblock Grounded-videollm: Sharpening fine-grained temporal grounding in video large language models.
\newblock \emph{arXiv preprint arXiv:2410.03290}, 2024{\natexlab{a}}.

\bibitem[Wang et~al.(2018)Wang, Xiong, Wang, Qiao, Lin, Tang, and Van~Gool]{wang2018temporal}
Limin Wang, Yuanjun Xiong, Zhe Wang, Yu Qiao, Dahua Lin, Xiaoou Tang, and Luc Van~Gool.
\newblock Temporal segment networks for action recognition in videos.
\newblock \emph{IEEE transactions on pattern analysis and machine intelligence}, 41\penalty0 (11):\penalty0 2740--2755, 2018.

\bibitem[Wang et~al.(2024{\natexlab{b}})Wang, Bai, Tan, Wang, Fan, Bai, Chen, Liu, Wang, Ge, et~al.]{Qwen2-VL}
Peng Wang, Shuai Bai, Sinan Tan, Shijie Wang, Zhihao Fan, Jinze Bai, Keqin Chen, Xuejing Liu, Jialin Wang, Wenbin Ge, et~al.
\newblock Qwen2-vl: Enhancing vision-language model's perception of the world at any resolution.
\newblock \emph{arXiv preprint arXiv:2409.12191}, 2024{\natexlab{b}}.

\bibitem[Wang et~al.(2023)Wang, Cao, and Zhang]{wang2023visual}
Yizhe Wang, Congqi Cao, and Yanning Zhang.
\newblock Visual-semantic network: a visual and semantic enhanced model for gesture recognition.
\newblock \emph{Visual Intelligence}, 1\penalty0 (1):\penalty0 25, 2023.

\bibitem[Wang et~al.(2024{\natexlab{c}})Wang, Meng, Liang, Wang, Liu, and Zhao]{Hawkeye}
Yueqian Wang, Xiaojun Meng, Jianxin Liang, Yuxuan Wang, Qun Liu, and Dongyan Zhao.
\newblock Hawkeye: Training video-text llms for grounding text in videos.
\newblock \emph{arXiv preprint arXiv:2403.10228}, 2024{\natexlab{c}}.

\bibitem[Wei et~al.(2024)Wei, Xiong, Ren, Du, Zhang, and Chen]{OmniEdit}
Cong Wei, Zheyang Xiong, Weiming Ren, Xinrun Du, Ge Zhang, and Wenhu Chen.
\newblock Omniedit: Building image editing generalist models through specialist supervision, 2024.

\bibitem[Wu et~al.(2024{\natexlab{a}})Wu, Zhang, Xia, Li, Xia, Chang, Yu, Kim, Rossi, Zhang, et~al.]{vp_survey}
Junda Wu, Zhehao Zhang, Yu Xia, Xintong Li, Zhaoyang Xia, Aaron Chang, Tong Yu, Sungchul Kim, Ryan~A Rossi, Ruiyi Zhang, et~al.
\newblock Visual prompting in multimodal large language models: A survey.
\newblock \emph{arXiv preprint arXiv:2409.15310}, 2024{\natexlab{a}}.

\bibitem[Wu et~al.(2024{\natexlab{b}})Wu, Huang, and Wang]{DOrA}
Tung-Yu Wu, Sheng-Yu Huang, and Yu-Chiang~Frank Wang.
\newblock Dora: 3d visual grounding with order-aware referring.
\newblock \emph{arXiv preprint arXiv:2403.16539}, 2024{\natexlab{b}}.

\bibitem[Wu and Yang(2024)]{wu2024glance}
Yongliang Wu and Xu Yang.
\newblock A glance at in-context learning.
\newblock \emph{Frontiers of Computer Science}, 18\penalty0 (5):\penalty0 185347, 2024.

\bibitem[Wu et~al.(2024{\natexlab{c}})Wu, Wang, Tang, Wu, He, Ouyang, Wu, and Torr]{Dettoolchain}
Yixuan Wu, Yizhou Wang, Shixiang Tang, Wenhao Wu, Tong He, Wanli Ouyang, Jian Wu, and Philip Torr.
\newblock Dettoolchain: A new prompting paradigm to unleash detection ability of mllm.
\newblock \emph{arXiv preprint arXiv:2403.12488}, 2024{\natexlab{c}}.

\bibitem[Wu et~al.(2024{\natexlab{d}})Wu, Zhu, Cao, Lu, Li, Chi, Qiu, Su, Zheng, Wu, et~al.]{wu2024video}
Yongliang Wu, Wenbo Zhu, Jiawang Cao, Yi Lu, Bozheng Li, Weiheng Chi, Zihan Qiu, Lirian Su, Haolin Zheng, Jay Wu, et~al.
\newblock Video repurposing from user generated content: A large-scale dataset and benchmark.
\newblock \emph{arXiv preprint arXiv:2412.08879}, 2024{\natexlab{d}}.

\bibitem[Xu et~al.(2021)Xu, Huang, and Liu]{xu2021sutd}
Li Xu, He Huang, and Jun Liu.
\newblock Sutd-trafficqa: A question answering benchmark and an efficient network for video reasoning over traffic events.
\newblock In \emph{Proceedings of the IEEE/CVF conference on computer vision and pattern recognition}, pages 9878--9888, 2021.

\bibitem[Xu et~al.(2022)Xu, Qu, Kuen, Gu, and Liu]{xu2022meta}
Li Xu, Haoxuan Qu, Jason Kuen, Jiuxiang Gu, and Jun Liu.
\newblock Meta spatio-temporal debiasing for video scene graph generation.
\newblock In \emph{European Conference on Computer Vision}, pages 374--390. Springer, 2022.

\bibitem[Yang et~al.(2023{\natexlab{a}})Yang, Nagrani, Seo, Miech, Pont-Tuset, Laptev, Sivic, and Schmid]{vid2seq}
Antoine Yang, Arsha Nagrani, Paul~Hongsuck Seo, Antoine Miech, Jordi Pont-Tuset, Ivan Laptev, Josef Sivic, and Cordelia Schmid.
\newblock Vid2seq: Large-scale pretraining of a visual language model for dense video captioning.
\newblock In \emph{Proceedings of the IEEE/CVF Conference on Computer Vision and Pattern Recognition}, pages 10714--10726, 2023{\natexlab{a}}.

\bibitem[Yang et~al.(2023{\natexlab{b}})Yang, Zhang, Li, Zou, Li, and Gao]{SoM}
Jianwei Yang, Hao Zhang, Feng Li, Xueyan Zou, Chunyuan Li, and Jianfeng Gao.
\newblock Set-of-mark prompting unleashes extraordinary visual grounding in gpt-4v.
\newblock \emph{arXiv preprint arXiv:2310.11441}, 2023{\natexlab{b}}.

\bibitem[Yang et~al.(2024)Yang, Wu, Yang, Chen, and Geng]{yang2024exploring}
Xu Yang, Yongliang Wu, Mingzhuo Yang, Haokun Chen, and Xin Geng.
\newblock Exploring diverse in-context configurations for image captioning.
\newblock \emph{Advances in Neural Information Processing Systems}, 36, 2024.

\bibitem[Yao et~al.(2024)Yao, Zhang, Zhang, Liu, Chua, and Sun]{CPT}
Yuan Yao, Ao Zhang, Zhengyan Zhang, Zhiyuan Liu, Tat-Seng Chua, and Maosong Sun.
\newblock Cpt: Colorful prompt tuning for pre-trained vision-language models.
\newblock \emph{AI Open}, 5:\penalty0 30--38, 2024.

\bibitem[Ye and Ma(2021)]{ye2021visual}
Xinyu Ye and Jiayi Ma.
\newblock Visual place recognition via local affine preserving matching.
\newblock In \emph{2021 IEEE International Conference on Robotics and Automation (ICRA)}, pages 12954--12960. IEEE, 2021.

\bibitem[Ye and Ma(2022)]{ye2022neighborhood}
Xinyu Ye and Jiayi Ma.
\newblock Neighborhood manifold preserving matching for visual place recognition.
\newblock \emph{IEEE Transactions on Industrial Informatics}, 19\penalty0 (7):\penalty0 8127--8136, 2022.

\bibitem[Ye et~al.(2023)Ye, Yan, and Yan]{ye2023vqne}
Xinyu Ye, Ge Yan, and Junchi Yan.
\newblock Vqne: Variational quantum network embedding with application to network alignment.
\newblock In \emph{Proceedings of the 29th ACM SIGKDD Conference on Knowledge Discovery and Data Mining}, pages 3105--3115, 2023.

\bibitem[Yi et~al.(2024)Yi, He, Zhan, and Ye]{yi2024bridge}
Chao Yi, Yuhang He, De-Chuan Zhan, and Han-Jia Ye.
\newblock Bridge the modality and capability gaps in vision-language model selection.
\newblock \emph{Advances in Neural Information Processing Systems}, 37:\penalty0 34429--34452, 2024.

\bibitem[Yin et~al.(2024)Yin, Wang, Guo, Liu, Zhang, Zhao, Lian, and Chen]{yin2024dataset}
Mingjia Yin, Hao Wang, Wei Guo, Yong Liu, Suojuan Zhang, Sirui Zhao, Defu Lian, and Enhong Chen.
\newblock Dataset regeneration for sequential recommendation.
\newblock In \emph{Proceedings of the 30th ACM SIGKDD Conference on Knowledge Discovery and Data Mining}, pages 3954--3965, 2024.

\bibitem[Zala et~al.(2023)Zala, Cho, Kottur, Chen, Oguz, Mehdad, and Bansal]{zala2023hierarchical}
Abhay Zala, Jaemin Cho, Satwik Kottur, Xilun Chen, Barlas Oguz, Yashar Mehdad, and Mohit Bansal.
\newblock Hierarchical video-moment retrieval and step-captioning.
\newblock In \emph{Proceedings of the IEEE/CVF Conference on Computer Vision and Pattern Recognition}, pages 23056--23065, 2023.

\bibitem[Zhang et~al.(2024{\natexlab{a}})Zhang, Zhang, Li, Zeng, Yang, Zhang, Wang, Tan, Li, and Liu]{LongVA}
Peiyuan Zhang, Kaichen Zhang, Bo Li, Guangtao Zeng, Jingkang Yang, Yuanhan Zhang, Ziyue Wang, Haoran Tan, Chunyuan Li, and Ziwei Liu.
\newblock Long context transfer from language to vision.
\newblock \emph{arXiv preprint arXiv:2406.16852}, 2024{\natexlab{a}}.

\bibitem[Zhang et~al.(2023)Zhang, Deng, Jia, Yu, Chen, Ma, Ding, and Zhang]{zhang2023pixel}
Wenyu Zhang, Xin Deng, Baojun Jia, Xingtong Yu, Yifan Chen, Jin Ma, Qing Ding, and Xinming Zhang.
\newblock Pixel adapter: A graph-based post-processing approach for scene text image super-resolution.
\newblock In \emph{Proceedings of the 31st ACM International Conference on Multimedia}, pages 2168--2179, 2023.

\bibitem[Zhang et~al.(2024{\natexlab{b}})Zhang, Wu, Li, Li, Ma, Liu, and Li]{LLaVA-Video}
Yuanhan Zhang, Jinming Wu, Wei Li, Bo Li, Zejun Ma, Ziwei Liu, and Chunyuan Li.
\newblock Video instruction tuning with synthetic data.
\newblock \emph{arXiv preprint arXiv:2410.02713}, 2024{\natexlab{b}}.

\bibitem[Zhao et~al.(2017)Zhao, Xiong, Wang, Wu, Tang, and Lin]{zhao2017temporal}
Yue Zhao, Yuanjun Xiong, Limin Wang, Zhirong Wu, Xiaoou Tang, and Dahua Lin.
\newblock Temporal action detection with structured segment networks.
\newblock In \emph{Proceedings of the IEEE international conference on computer vision}, pages 2914--2923, 2017.

\bibitem[Zhao et~al.(2024)Zhao, Tang, Lin, Wu, Huang, Liu, Tan, Zhang, and Xie]{zhao2024multi}
Zhen Zhao, Jingqun Tang, Chunhui Lin, Binghong Wu, Can Huang, Hao Liu, Xin Tan, Zhizhong Zhang, and Yuan Xie.
\newblock Multi-modal in-context learning makes an ego-evolving scene text recognizer.
\newblock In \emph{Proceedings of the IEEE/CVF Conference on Computer Vision and Pattern Recognition}, pages 15567--15576, 2024.

\bibitem[Zhou et~al.(2024)Zhou, Qin, Yin, Huang, Zhang, Sheng, Qiao, and Shao]{Minedreamer}
Enshen Zhou, Yiran Qin, Zhenfei Yin, Yuzhou Huang, Ruimao Zhang, Lu Sheng, Yu Qiao, and Jing Shao.
\newblock Minedreamer: Learning to follow instructions via chain-of-imagination for simulated-world control.
\newblock \emph{arXiv preprint arXiv:2403.12037}, 2024.

\end{thebibliography}
}

\clearpage
\setcounter{page}{1}
\maketitlesupplementary
\section{Experimental Details}
\subsection{Moment Retrieval} In the training-free (NumPro) setting, we extract frames from videos at 1 FPS, with each frame resized to a resolution of \(336 \times 336\). In the fine-tuned (NumPro-FT) setting, frames are extracted at 0.5 FPS during both the training and inference phases due to GPU memory constraints. 

\subsection{Highlight Detection} In both the training-free (NumPro) and fine-tuned (NumPro-FT) settings, we extract frames from videos at 0.5 FPS because the saliency score is labeled every 2 seconds. Each frame is resized to a resolution of \(336 \times 336\).

\section{Hallucination in Vid-LLMs for VTG}
\subsection{General Vid-LLMs}
\label{sec:general_vid}
\begin{figure*}[t]
    \centering
    \begin{subfigure}[t]{0.49\linewidth}
        \centering
        \includegraphics[width=\linewidth]{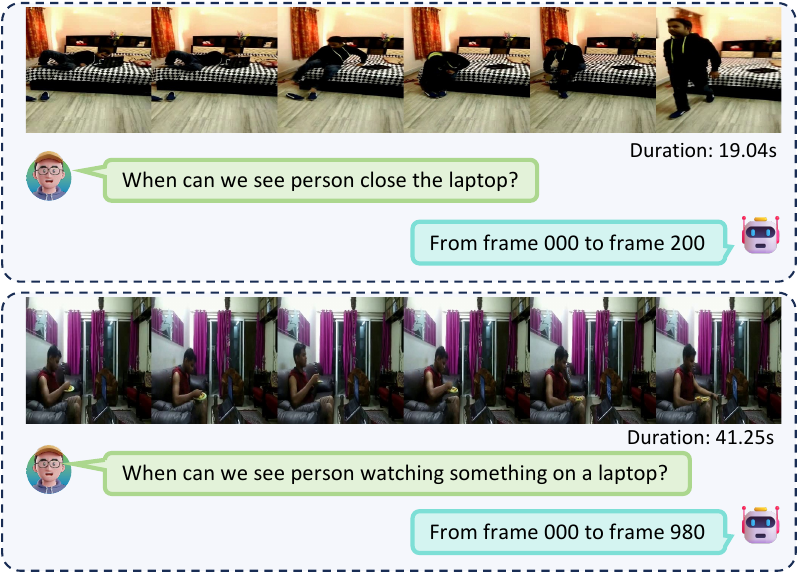}
        \caption{Qwen2-VL-7B}
        \label{fig:qwen2_bad}
    \end{subfigure}
    \hfill
    \begin{subfigure}[t]{0.49\linewidth}
        \centering
        \includegraphics[width=\linewidth]{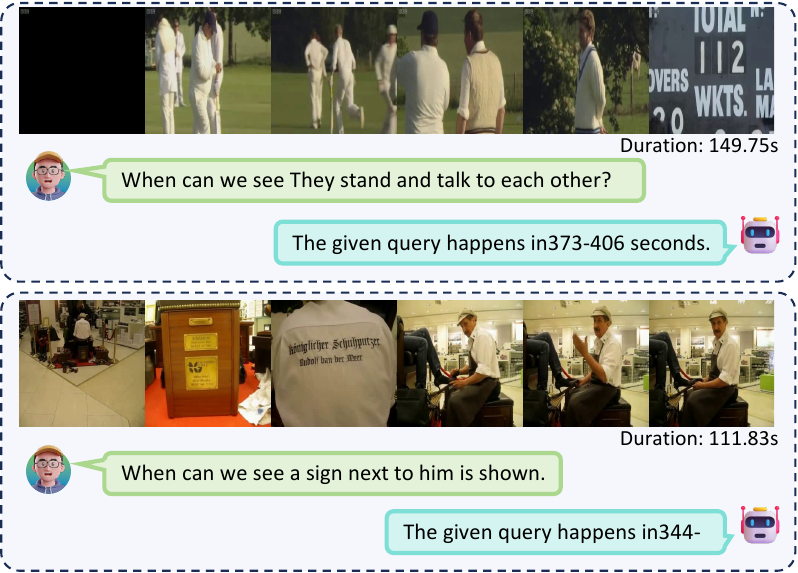}
        \caption{Qwen2-VL-72B}
        \label{fig:qwen72b_bad}
    \end{subfigure}
    \caption{Video temporal grounding results where the models exhibit serious hallucination and output incorrect results. In all cases, frames are sampled at 1 FPS.}
    \label{fig:combined_bad}
\end{figure*}

\noindent \textbf{Qwen2-VL-7B.} Figure~\ref{fig:qwen2_bad} shows the results of Qwen2-VL-7B~\cite{Qwen2-VL} suffer from severe hallucinations. For instance, the model generates responses like ``from frame 000 to frame 200" even when the input video contains only 19 frames.

\noindent \textbf{Qwen2-VL-72B.} A larger-scale Vid-LLM, Qwen2-VL-72B~\cite{Qwen2-VL}, as shown in Figure~\ref{fig:qwen72b_bad}, also exhibits significant hallucination issues. The model produces illogical outputs, such as incomplete sentences like “The given query happens in344-,” further emphasizing its struggle with coherent and accurate temporal reasoning.

\noindent \textbf{LLaVA-Video-7B.} Figure~\ref{fig:llava_video} displays the distribution of the top 10 most common time intervals predicted by LLaVA-Video-7B~\cite{LLaVA-Video}. The model frequently outputs very short segments, such as $[1,3]$, $[2,4]$, and $[2,3]$, which together account for over 50\% of predictions. This behavior suggests a significant bias in the model toward producing overly simplistic temporal spans.

\noindent \textbf{LLaVA-OneVision-7B.} Figure~\ref{fig:llava_onevision} presents the 10 most frequently predicted intervals of LLaVA-OneVision-7B~\cite{LLaVA-OneVision}. The outputs are dominated by segments like $[10,12]$ and $[1,3]$, which together account for over 50\% of all results. The repetitive predictions indicate a substantial hallucination problem limiting temporal reasoning capability.

\subsection{VTG-Tuned Vid-LLMs}
\label{sec:vtg_vid}
\begin{figure*}[t]
    \centering
    \begin{subfigure}[t]{0.49\linewidth}
        \centering
        \includegraphics[width=\linewidth]{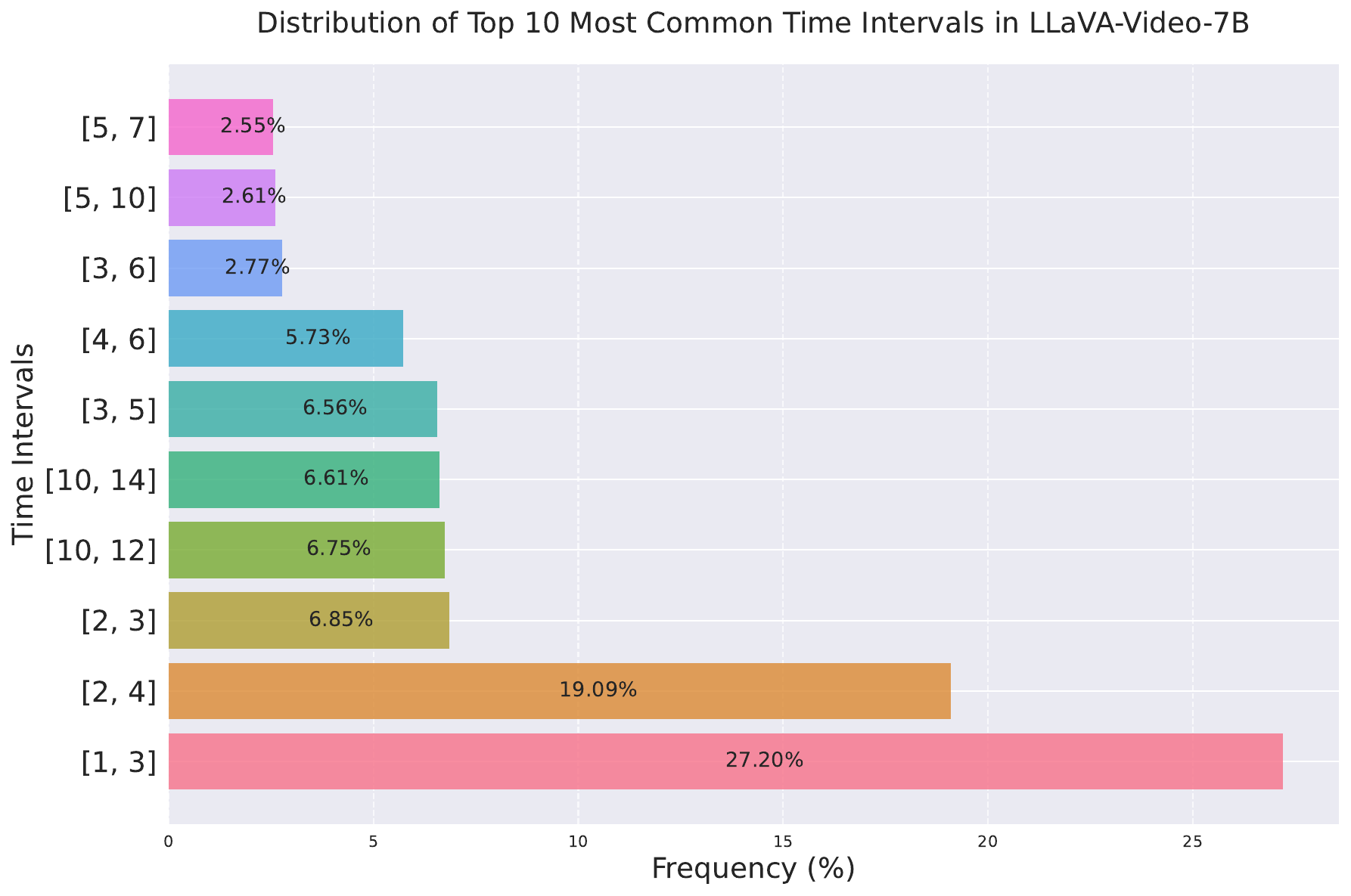}
        \caption{LLaVA-Video-7B}
        \label{fig:llava_video}
    \end{subfigure}
    \hfill
    \begin{subfigure}[t]{0.49\linewidth}
        \centering
        \includegraphics[width=\linewidth]{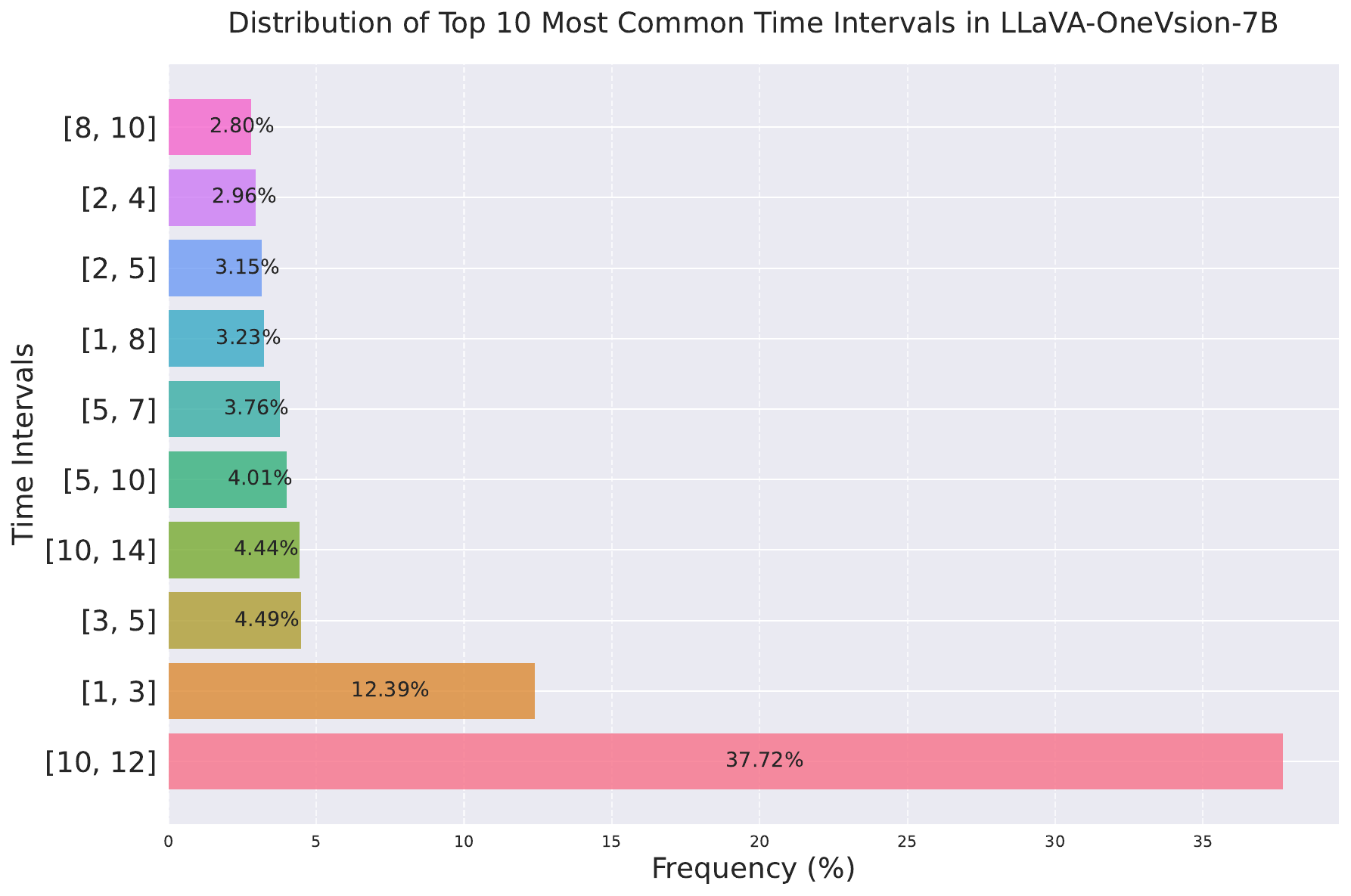}
        \caption{LLaVA-OneVision-7B}
        \label{fig:llava_onevision}
    \end{subfigure}
    
    \begin{subfigure}[t]{0.49\linewidth}
        \centering
        \includegraphics[width=\linewidth]{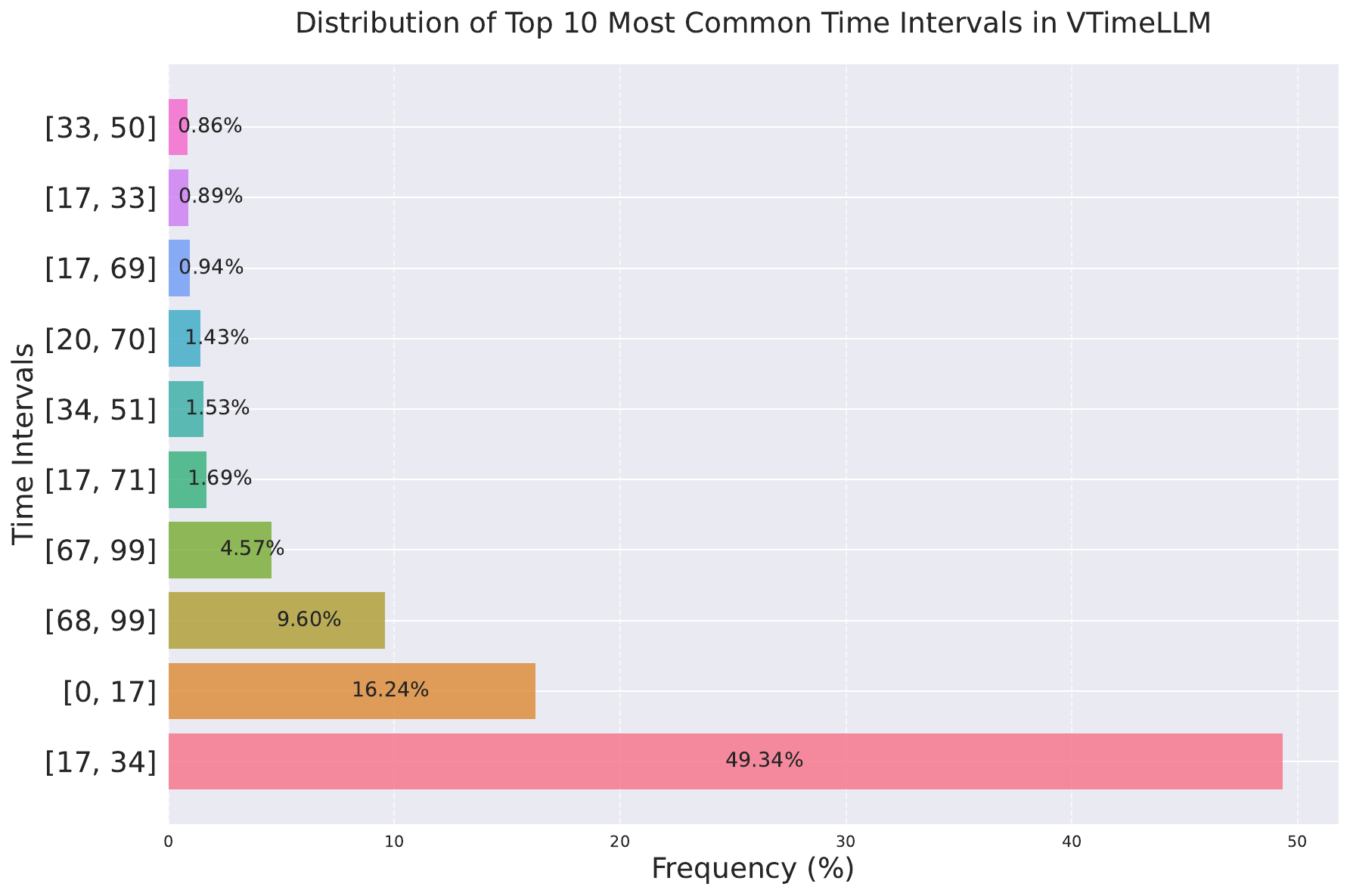}
        \caption{VTimeLLM}
        \label{fig:vtime_charades}
    \end{subfigure}
    \hfill
    \begin{subfigure}[t]{0.49\linewidth}
        \centering
        \includegraphics[width=\linewidth]{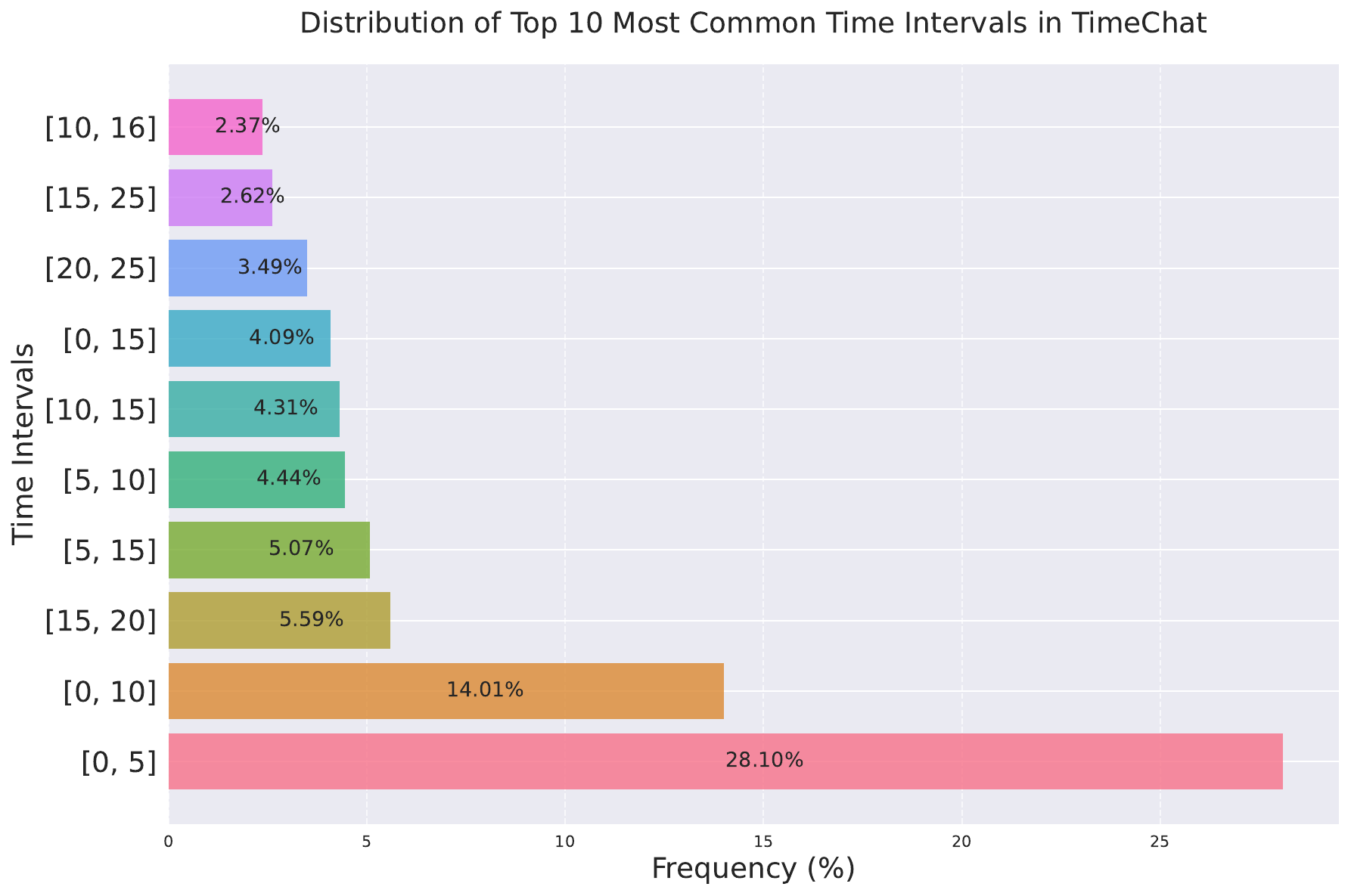}
        \caption{TimeChat}
        \label{fig:timechat_charades}
    \end{subfigure}
    \caption{Distribution of the Top 10 Most Common Time Intervals within the Charades-STA Dataset for Different Models.}
    \label{fig:combined}
\end{figure*}

\noindent \textbf{VTimeLLM.} Figure~\ref{fig:vtime_charades} shows the predictions on the Charades-STA dataset of VTimeLLM~\cite{VTimeLLM}. The model frequently predicts certain frame intervals, such as $[17,34]$, which accounts for 49.34\% of predictions, and $[0,17]$, which constitutes 16.34\%. This pattern suggests significant hallucination and overfitting to specific frame numbers.

\noindent \textbf{TimeChat.} As shown in Figure~\ref{fig:timechat_charades}, we analyze the output of the TimeChat~\cite{TimeChat} model. It tends to produce results in multiples of 5, such as 5, 10, 15, and 20. Notably, intervals like $[0,5]$ and $[0,10]$ appear in over 42\% of predictions. This indicates both hallucinations and overfitting.
\begin{table}[t]
\centering
\caption{The ablation results on the QVHighlights dataset.}
\vspace{-10pt}
\begin{tabular}{@{}l|ll@{}}
\toprule
\multirow{2}{*}{Model} &  \multicolumn{2}{c}{QVHighlights}\\
 & mAP & HIT@1\\ 
\midrule
LLaVA-OneVision-7B & 17.2 & 19.9 \\
\rowcolor{gray!20} ~~~~~~\textit{+NumPro} & $20.9_\impro{3.7}$ & $27.6_\impro{7.7}$ \\
\midrule
LLaVA-Video-7B & 20.7  & 34.8 \\
\rowcolor{gray!20} ~~~~~~\textit{+NumPro} & $22.3_\impro{1.6}$ & $38.4_\impro{4.4}$ \\
\midrule
Qwen2-VL-72B & 21.6 &  37.5 \\
\rowcolor{gray!20} ~~~~~~\textit{+NumPro} & $24.2_\impro{2.6}$ & $44.3_\impro{6.8}$ \\
\midrule
LongVA-7B-DPO & 14.2 & 20.4 \\
~~~~~~\textit{+FT} & 21.9 & 30.8 \\
\rowcolor{gray!20} ~~~~~~\textit{+NumPro-FT} & $25.0_\impro{10.8}$ & $37.2_\impro{16.8}$ \\
\bottomrule
\end{tabular}
\label{table:ablation_qvh}
\end{table}

\begin{table}[t]
\centering
\caption{Ablation study with different font size of NumPro-FT.}
\vspace{-10pt}
\resizebox{\linewidth}{!}{
\begin{tabular}{@{}lll|cccc@{}}
\toprule
\multirow{2}{*}{Size} & \multirow{2}{*}{Color} & \multirow{2}{*}{Position} & \multicolumn{4}{c}{Charades-STA}  \\
 &  &  & R@0.3 & R@0.5 & R@0.7 & mIoU \\ 
 \midrule
40 & Red & Bottom Right & \textbf{63.8} & \textbf{42.0} & \textbf{20.6} & \textbf{41.4} \\
60 & Red & Bottom Right & 56.0 & 37.6 & \textbf{20.6} & 40.9 \\
\bottomrule
\end{tabular}
}
\label{table:ablation_design_longva}
\end{table}

\section{Additional Attention Analysis}
\label{sec:attn_cases}

\begin{table}[t]
\centering
\caption{Performance comparison between the original Qwen2-VL-7B and the ``Attention Map" method, which selects the two frames with the highest attention scores as the temporal boundaries. }
\vspace{-10pt}
\begin{tabular}{@{}l|cccc@{}}
\toprule
\multirow{2}{*}{Method} & \multicolumn{4}{c}{Charades-STA}  \\
 & R@0.3 & R@0.5 & R@0.7 & mIoU \\ 
 \midrule
Qwen2-VL-7B & 8.7 & 5.4 & 2.4 & 7.9  \\
Attention Map & \textbf{18.1} & \textbf{11.4} & \textbf{3.1} & \textbf{19.8} \\
\bottomrule
\end{tabular}

\label{table:attn_results}
\end{table}

\begin{figure*}[t]
    \centering
    \includegraphics[width=1\linewidth]{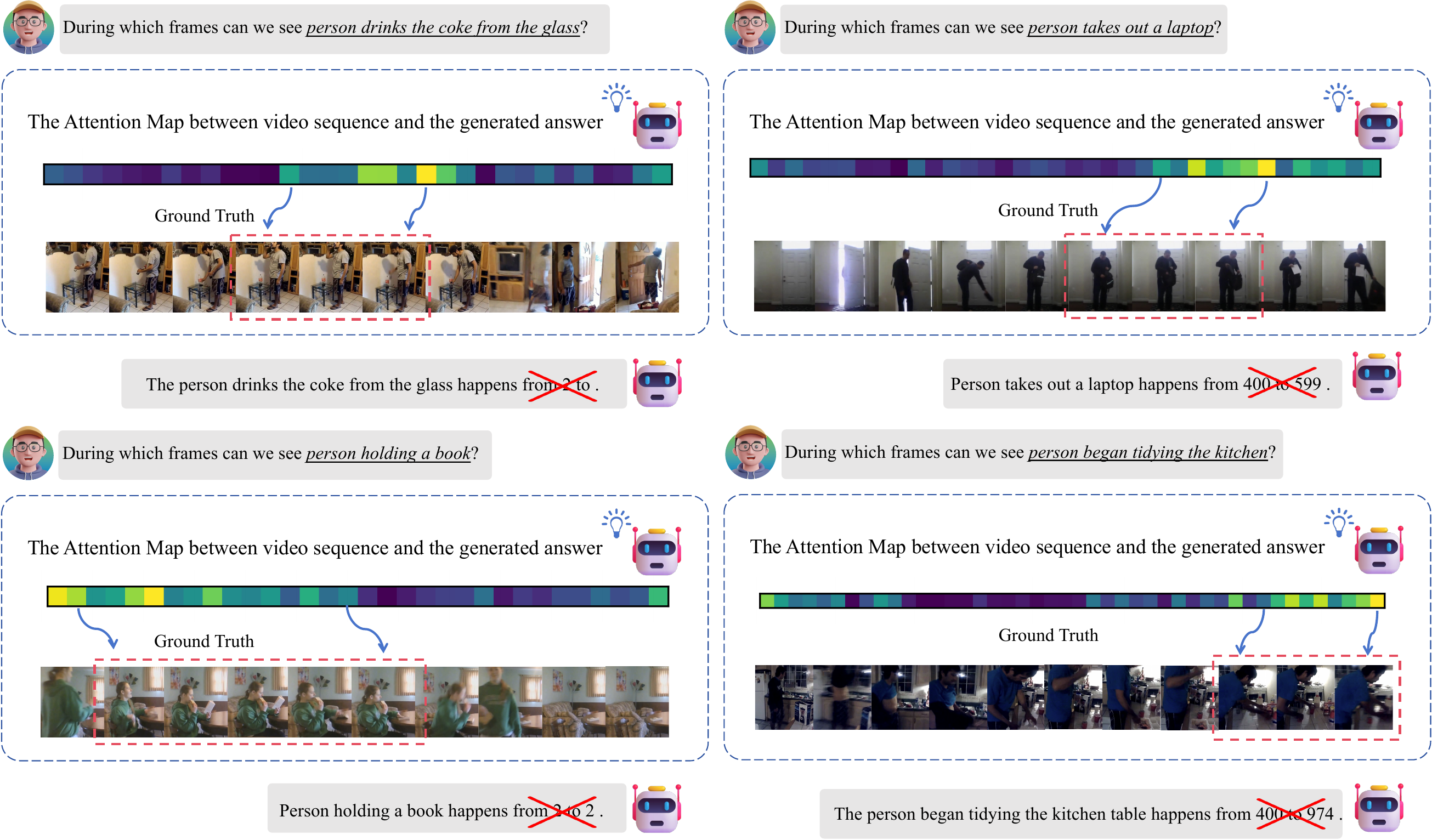}
    \caption{Additional attention analysis results on general Vid-LLMs}
    \label{fig:attn_cases}
\end{figure*}

We present additional attention analysis results in Figure~\ref{fig:attn_cases}. In the examples on the left, the model produces incorrect or incomplete outputs, such as ``from 2 to .". On the right, the examples display severe hallucinations, with outputs extending beyond the video's actual duration. Despite these issues, the attention maps in both cases consistently highlight the relevant video segments. 
These findings show that while Vid-LLMs identify correct video segments, they fail to output accurate temporal boundaries due to the inability to translate these segments into precise textual locations.

To further examine this challenge of Vid-LLMs quantitatively, we conduct an experiment with the Qwen2-VL-7B model on the Charades-STA dataset. In this experiment, the two frames with the highest attention scores are selected as the predicted start and end frames of the segment. Specifically, we selected the two frames with the highest attention scores as the start and end frames, treating these as the predicted segment boundaries. The results, presented in Table~\ref{table:attn_results}, showing that this na\"ive attention-based solution achieves an improvement of 11.9\% in mIoU compared to direct predictions of the original model. This significant gain supports our observation that Vid-LLMs have the inherent capacity to locate relevant video segments but struggle to express temporal boundaries accurately in text.

Overall, these analyses highlight the primary bottleneck of Vid-LLMs in addressing temporal grounding tasks and emphasize the need to overcome this limitation, which is what our proposed NumPro method is designed to address.
% These findings support our hypothesis that, while the model can identify the correct segments, it fails to generate accurate grounding outputs due to the absence of explicit timestamp information.

% To further illustrate the relevance of this attention mechanism, we employ the Qwen2-VL-7B model and conduct quantitative experiments on the Charades-STA dataset. We identify the two frames with the highest scores on the attention map, designating them as the start and end frames. The results are presented in  It is evident that utilizing the attention-based approach (as shown in the attention map in the table) significantly outperformed direct predictions from the model, achieving an improvement of +11.9 in mIoU.

\section{Additional Video Benchmark Results}
\label{sec:more_results_videoqa}
We conducted experiments on additional video question answering (QA) benchmarks, MVBench~\cite{Mvbench} and VideoMME~\cite{Video-MME}, as summarized in Table~\ref{tab:more_results_videoqa}. We use 1FPS as the sampling rate, and adopt a design of red color, font size 40, and bottom right positioning for the number prompt. Our results demonstrate that Vid-LLMs enhanced with NumPro achieve robust performance across diverse downstream tasks. Notably, NumPro significantly improves the Vid-LLMs' generalization capabilities in temporal understanding tasks, such as Scene Transition and Temporal Perception. These findings align with our previous results presented in Table~\ref{tab:video_qa} on VideoInstruct~\cite{Video-ChatGPT} in the main paper.

\begin{table}[h]
\caption{Evaluations on two video QA benchmarks: MVBench and VideoMME. The results demonstrate that our NumPro approach enhances Vid-LLMs' generalization capabilities on downstream tasks involving temporal understanding.}
\centering
\vspace{-10pt}
\resizebox{\linewidth}{!}{
\begin{tabular}{@{}ccc@{\hskip 1pt}ccc@{}}
\toprule
\multicolumn{3}{c}{\textbf{MVBench}} & \multicolumn{3}{c}{\textbf{Video-MME}} \\ \cmidrule(lr){1-3} \cmidrule(lr){4-6}
\textbf{Scene Transition} & \textbf{State Change} & \textbf{Overall} &
\textbf{Temp. Per.} & \textbf{Temp. Rea.} & \textbf{Overall} \\ \midrule
$80.0_\impro{2.5}$      & $42.0_\impro{1.0}$   & $51.8_\impro{0.2}$ &
$72.7_\impro{12.7}$     & $49.7_\impro{8.8}$   & $63.7_\impro{0.3}$ \\ \bottomrule
\end{tabular}}
\vspace{-13pt}
\label{tab:more_results_videoqa}
\end{table}

\section{Ablation Results on Highlight Detection}
\label{sec:ablation_qvh}
We present additional ablation results on the QVHighlights dataset, as shown in Table~\ref{table:ablation_qvh}. The results show that our method generalizes well across various General Vid-LLMs, achieving notable improvements in both mAP and HIT@1 metrics. Specifically, fine-tuning with NumPro-FT obtains a 10.8\% increase in mAP and a 16.8\% increase in HIT@1, surpassing state-of-the-art results.

\section{Ablation Results on NumPro-FT Designs}
\label{sec:ablation_design_longva}
In this section, we present the ablation results for NumPro-FT. While a font size of 60 achieves better Number Accuracy than a size of 40 (Figure 5 of the main paper), it reduces Caption Accuracy and introduces more outliers. Table~\ref{table:ablation_design_longva} further supports this finding, showing that a font size of 60 results in generally lower performance, including a 7.8\% drop in R@0.3 compared to size 40. We attribute this to interference with the model's understanding of video content.

\begin{table}[h]
\centering
\caption{Comparison between overlaying timestamps with overlaying frame numbers in NumPro design.}
\vspace{-10pt}
\resizebox{\linewidth}{!}{
\begin{tabular}{@{}lcccc@{}} 
\toprule
Dataset   & R@0.3 & R@0.5 & R@0.7 & mIoU \\ 
\midrule
Charades-STA           & $58.4_\decli{2.3}$          & $37.8_\impro{1.0}$          & $16.6_\impro{0.7}$          & $37.6_\decli{0.9}$         \\
ActivityNet        & $31.6_\decli{12.6}$          & $18.1_\decli{6.3}$          & $10.2_\decli{4.2}$          & $23.0_\decli{8.3}$         \\ \bottomrule
\end{tabular}}
\vspace{-13pt}
\label{tab:timestamp_design}
\end{table}

\section{NumPro with Accurate Timestamps}
\label{sec:timestamp_design}
In the main paper, we choose frame numbers because they serve as the smallest discrete units of a video and can be directly mapped to precise timestamps using the frame sampling rate. In this section, we compare the performance by directly overlaying actual timestamps in videos.  
However, timestamps (\textit{e.g.}, ``10.5s") may introduce decimals, which can increase parsing complexity for Vid-LLMs. In Tabel~\ref{tab:timestamp_design}, we compare minute-level temporal annotations (\textit{e.g.}, ``01:10") with frame numbers annotations sampled with 1FPS. The performance on Charades-STA is comparable, while frame numbers outperformed timestamps on ActivityNet, which includes longer videos and more annotations. These suggest that overlaying numbers with temporal information is an effective strategy for Vid-LLMs in temporal grounding, while frame numbers offer a simpler and more scalable solution.

\section{Additional Visualization Cases}
\label{sec:more_cases}
\subsection{Dialogue} 
\begin{figure*}[t]
    \centering
    \includegraphics[width=\linewidth]{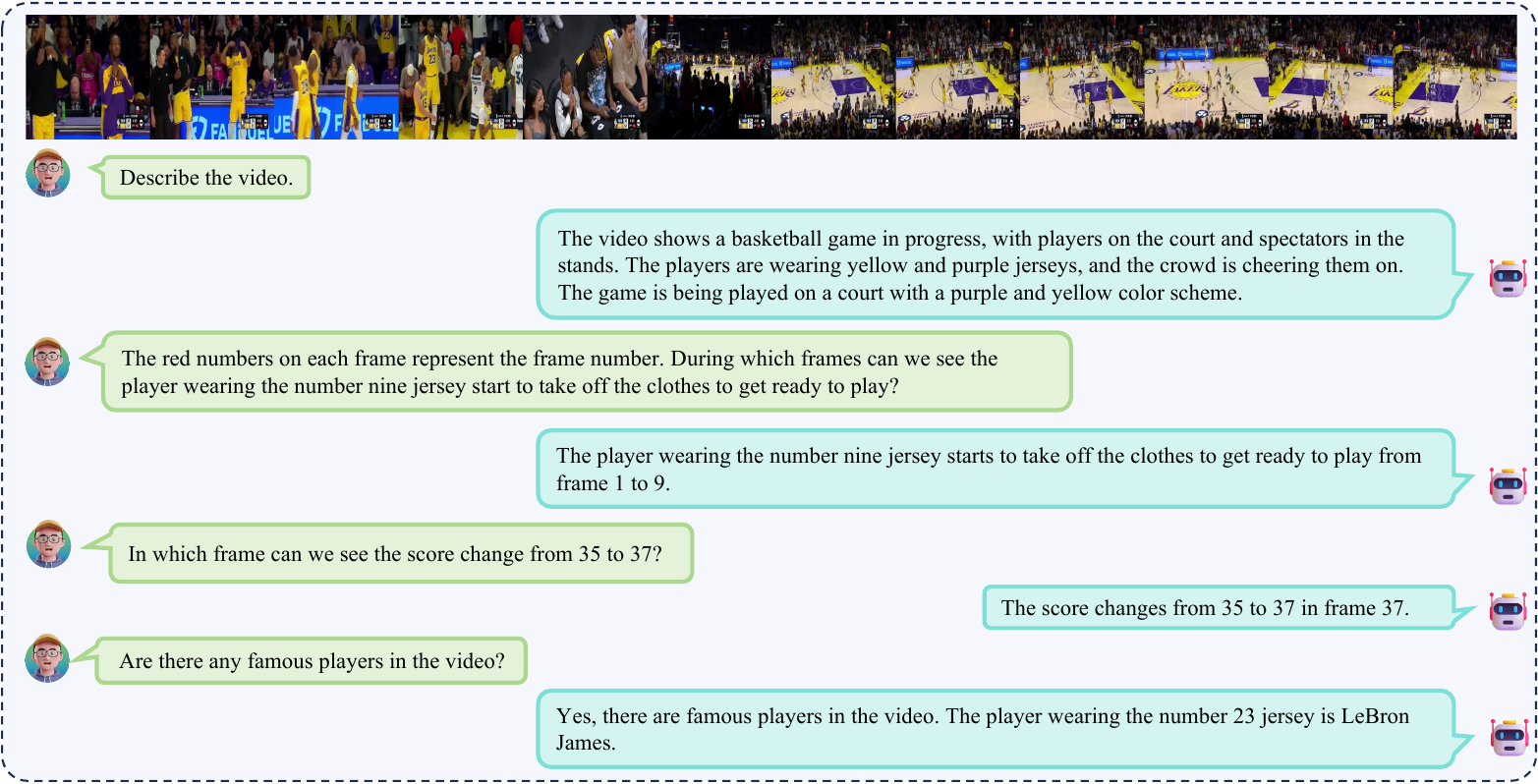}
    \caption{An example of multi-turn dialogue using our NumPro method in the Qwen2-VL-7B model.}
    \label{fig:dialogue}
\end{figure*}

Figure~\ref{fig:dialogue} illustrates a real-world application of our NumPro method within the Qwen2-VL-7B model, highlighting its ability to handle complex video-based dialogue tasks. Compared to the VTG-specific models~\cite{chen2021end,lin2023univtg}, NumPro equipped with Vid-LLMs facilitates multi-turn dialogue that adapt to user queries in real-time. For instance, the model can track score changes across video segments, identify celebrities through advanced facial recognition, and even extract textual information via OCR. These capabilities demonstrate the enhanced contextual understanding and practical value of Vid-LLMs for video comprehension tasks. By integrating NumPro, our approach further refines the temporal grounding process, enabling more precise and interactive video analysis for real-world applications.

\subsection{Moment Retrieval} 
\begin{figure*}[t]
    \centering
    \includegraphics[width=\linewidth]{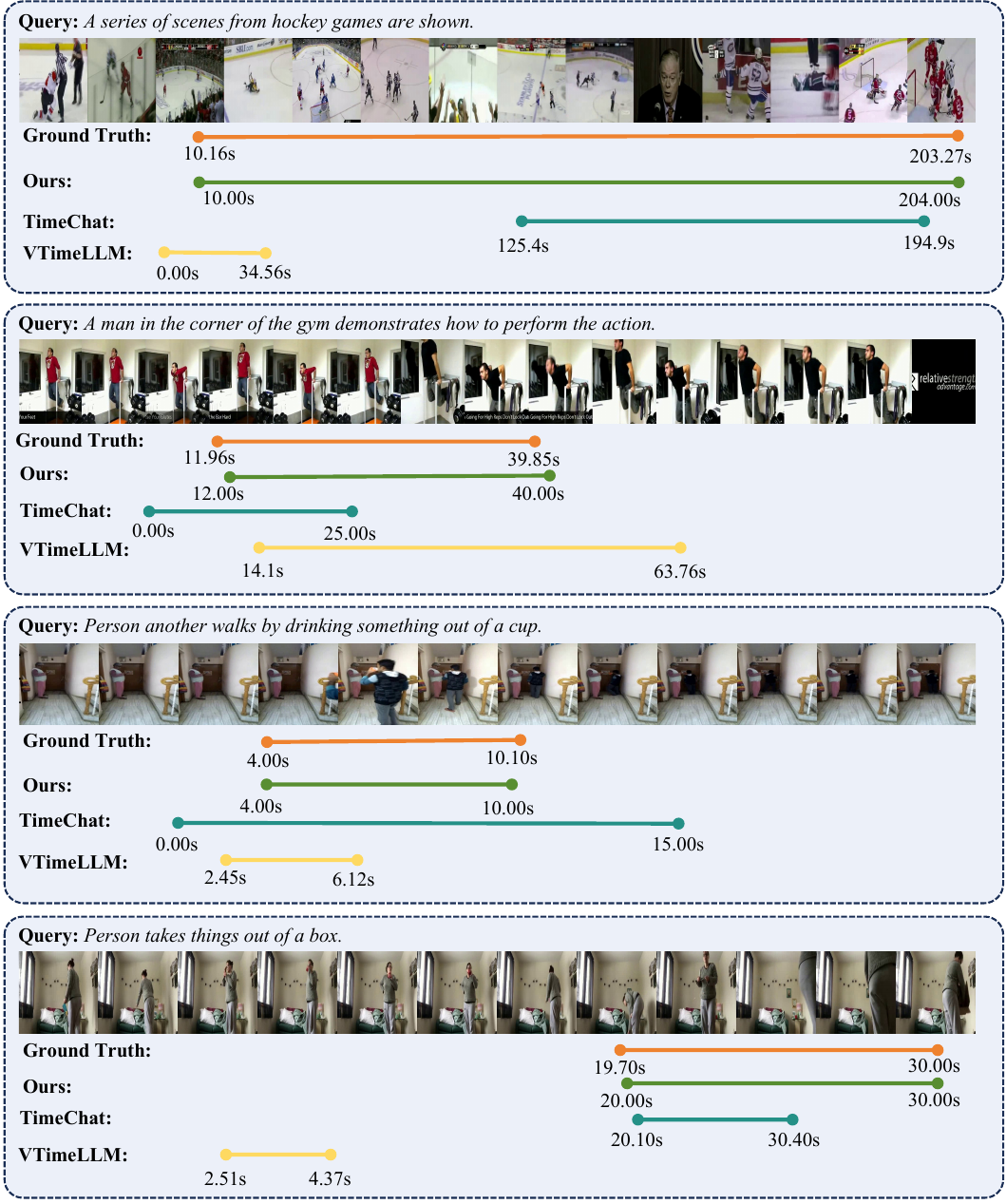}
    \caption{Additional visualization cases of Video Temporal Grounding task.}
    \label{fig:vtg_cases}
\end{figure*}

Figure~\ref{fig:vtg_cases} showcases additional visualization examples highlighting the effectiveness of our method in moment retrieval tasks. Our approach demonstrates robust temporal grounding capabilities by accurately identifying event boundaries across videos of varying lengths and content. Compared to previous state-of-the-art methods, it achieves substantial performance improvements.

\subsection{Highlight Detection} 
\begin{figure*}[t]
    \centering
    \includegraphics[width=\linewidth]{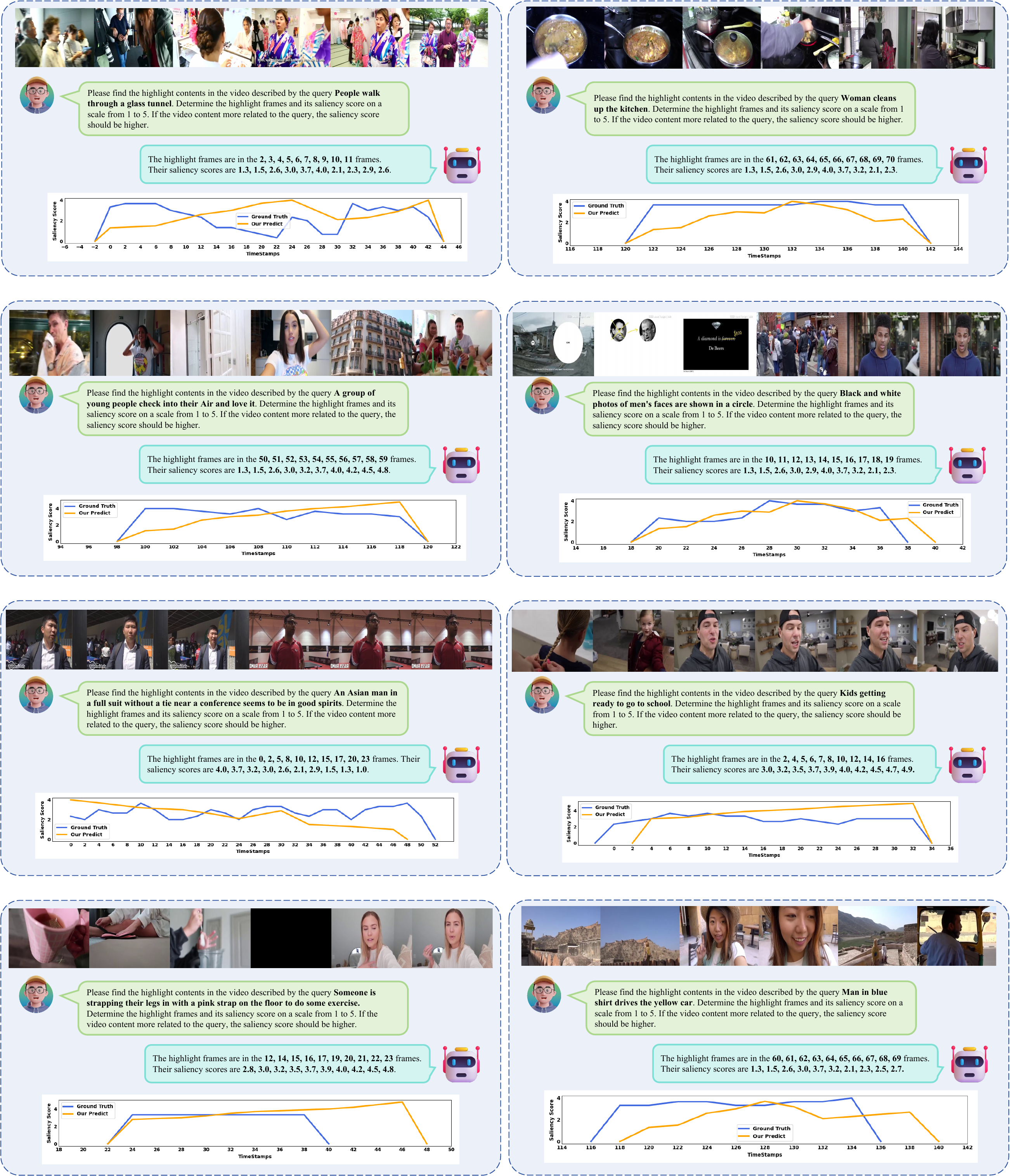}
    \caption{Additional visualization cases of Highlight Detection task on QVHighlights dataset.}
    \label{fig:qvh_cases}
\end{figure*}

Highlight detection~\cite{QVHighlights} focuses on identifying video segments that match a given query while also assessing their relative importance. The model generates timestamps for the relevant segments and assigns a saliency score on a scale from 1 to 5. As shown in Figure~\ref{fig:qvh_cases}, our method excels in accurately predicting segment start and end times, achieving consistently high mAP values. Additionally, it demonstrates precision in saliency score assessment, highlighting its suitability for tasks requiring detailed temporal localization and importance evaluation.

\section{Limitations}
While NumPro and NumPro-FT have proven effective across multiple models and datasets, significantly surpassing previous state-of-the-art models, there are still some limitations:
\begin{itemize}
    \item Limited Dataset Scope: Current datasets for video temporal grounding (VTG) tasks are predominantly focused on short videos, typically ranging from 30 seconds to 3 minutes in duration. Expanding evaluation to include longer videos, such as hour-long recordings, is essential for testing the scalability and generalizability of our approach.
    \item Potential Visual Obstruction: Although NumPro is designed to minimize its impact on video content, there are scenarios where it might obscure critical visual elements, such as details, watermarks, or logos. Future enhancements could involve dynamic adjustments to the opacity of numbers or the implementation of adaptive number positioning to avoid blocking essential visual information.
    \item Frame Rate Optimization: The effect of different sampling frame rates on performance remains underexplored. This study used a fixed frame rate of 1 FPS for NumPro, which may not be universally optimal. Investigating adaptive frame rates that align with the perceptual and computational characteristics of various models could lead to further improvements in accuracy and efficiency.
\end{itemize}

% These areas are important directions for future exploration.

\end{document}